\documentclass[iicol]{sn-jnl}

\jyear{2021}%

\theoremstyle{thmstyleone}%
\theoremstyle{thmstyletwo}%

\theoremstyle{thmstylethree}%

\usepackage[utf8]{inputenc} 
\usepackage[T1]{fontenc}    
\usepackage{hyperref}       
\usepackage{url}            
\usepackage{booktabs}       
\usepackage{amsfonts}       
\usepackage{nicefrac}       
\usepackage{microtype}      
\usepackage{xcolor}         
\usepackage{bbding}
\usepackage{color}
\usepackage{graphicx}
\usepackage{wrapfig}
\usepackage{subcaption}
\usepackage{multirow}
\usepackage{bbding, pifont}
\usepackage{amsmath,amssymb,eucal} 
\usepackage{ragged2e}
\usepackage{arydshln}
\usepackage[authoryear,sort&compress]{natbib}

\begin{document}

\title[Super Vision Transformer]{Super Vision Transformer}


\author[1,2]{\fnm{Mingbao} \sur{Lin}}
\equalcont{These authors contributed equally to this work.}

\author[1]{\fnm{Mengzhao} \sur{Chen}}
\equalcont{These authors contributed equally to this work.}

\author[1]{\fnm{Yuxin} \sur{Zhang}}

\author[3]{\fnm{Chunhua} \sur{Shen}}
\author[1]{\fnm{Rongrong} \sur{Ji}}
\author*[1]{\fnm{Liujuan} \sur{Cao}}\email{caoliujuan@xmu.edu.cn}

\affil[1]{\orgdiv{Key Laboratory of Multimedia Trusted Perception and Efficient Computing, Ministry of Education of China}, \orgname{School of Informatics}, \orgaddress{\city{Xiamen}, \country{China}}}

\affil[2]{\orgdiv{Tencent Youtu Lab},  \orgaddress{\city{Shanghai}, \country{China}}}

\affil[3]{\orgname{Zhejiang University}, \orgaddress{\city{Hangzhou},  \country{China}}}


\abstract{We attempt to reduce the computational costs in vision transformers (ViTs), which increase quadratically in the token number. We present a novel training paradigm that trains only one ViT model at a time, but is capable of providing improved image recognition performance with various computational costs. Here, the trained ViT model, termed super vision transformer (SuperViT), is empowered with the versatile ability to solve incoming patches of multiple sizes as well as preserve informative tokens with multiple keeping rates (the ratio of keeping tokens) to achieve good hardware efficiency for inference, given that the available hardware resources often change from time to time. Experimental results on ImageNet demonstrate that our SuperViT can considerably reduce the computational costs of ViT models with even performance increase. For example, we reduce 2$\times$ FLOPs of DeiT-S while increasing the Top-1 accuracy by 0.2\% and 0.7\% for 1.5$\times$ reduction. Also, our SuperViT significantly outperforms existing studies on efficient vision transformers. For example, when consuming the same amount of FLOPs, our SuperViT surpasses the recent state-of-the-art (SOTA) EViT by 1.1\% when using DeiT-S as their backbones. The project of this work is made publicly available at \url{https://github.com/lmbxmu/SuperViT}.}




\maketitle

\section{Introduction}\label{introduction}
Vision transformers (ViTs) initially introduced in 2020~\citep{dosovitskiy2020image} have spread widely in the field of computer vision and soon become one of the most pervasive and promising architectures in varieties of prevalent vision tasks, such as image classification~\citep{dosovitskiy2020image, jiang2021all,graham2021levit}, object detection~\citep{carion2020end,zhu2020deformable}, video understanding~\citep{bertasius2021space,arnab2021vivit} and many others~\citep{zheng2021rethinking,xie2021segformer,liang2021swinir,zamir2021restormer,huang2020hand}.
The basic idea behind ViTs is to break down an image as a series of local patches and use a linear projection to tokenize these patches as inputs.
In particular, ViTs merit in its property of capturing the long-range relationships between different portions of an image with the mechanism of multi-head self-attention (MHSA). Therefore, increasing attention has been paid to developing ViTs in various vision tasks.
%

\begin{figure*}[!t]
\begin{center}
\includegraphics[width=\linewidth]{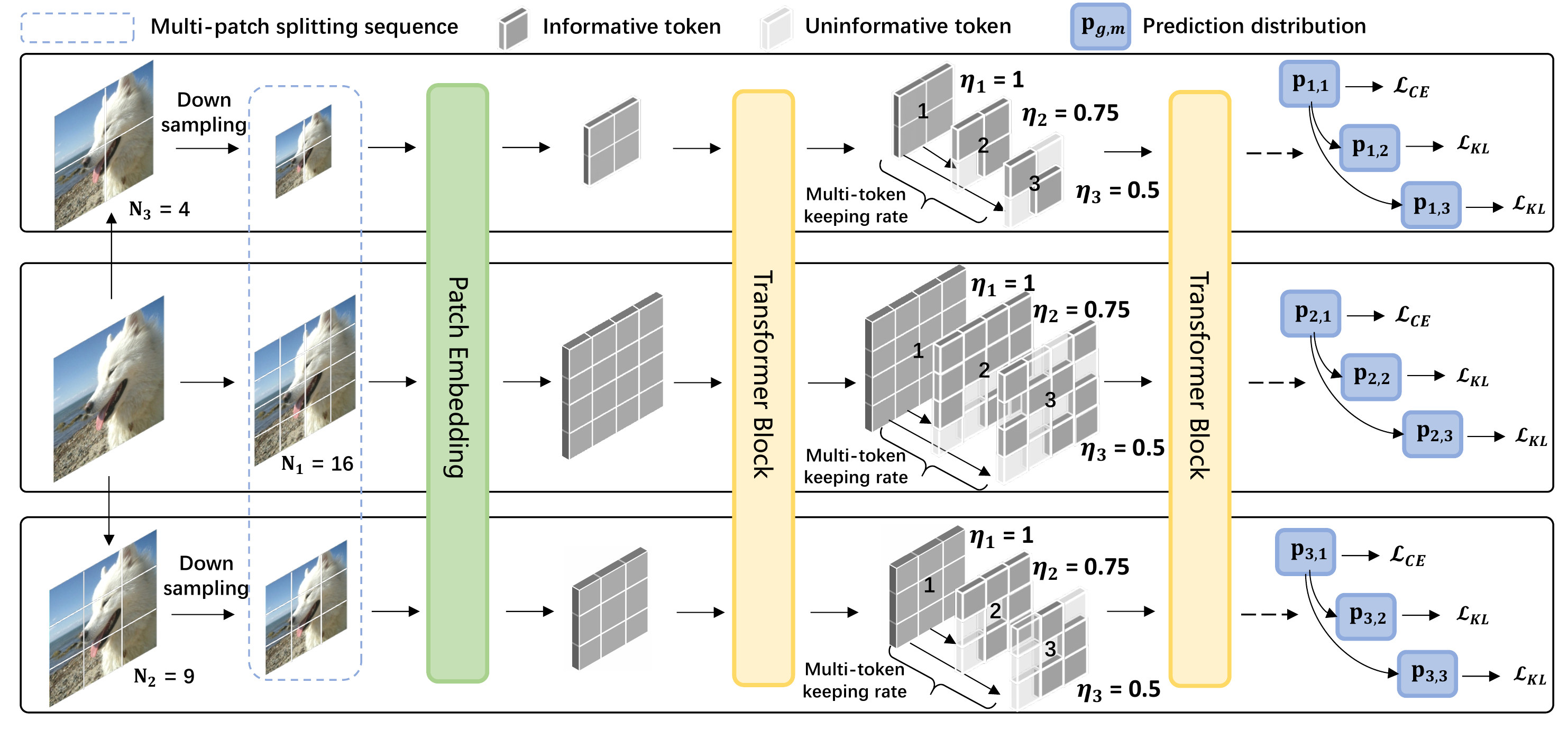}
\end{center}
\caption{\label{framework}Training framework of SuperViT. An image is arranged into multiple branches of different patch sizes. For each branch, various token keeping rates are considered in ViT training. Thus, the trained SuperViT is capable of better recognizing images with different computational costs.}
\end{figure*}

Recent studies focus more on an efficient ViT~\citep{liu2021swin,chu2021twins,li2022sepvit,graham2021levit,chavan2022vision} since the excessive computational costs, which increase quadratically to the number of tokens, have severely barricaded the broader usage of ViTs in real-world applications.
Note that the transformer's token sequence length is inversely proportional to the square of the patch size, which denotes that models with smaller patch sizes are computationally more expensive.
The most intuitive way is to reduce the transformer's token number by enlarging the patch size.
However, it has been an experimental consensus in the literature~\citep{dosovitskiy2020image} that a ViT model performs better with smaller-size patches as its inputs. For example, ViT-B~\citep{dosovitskiy2020image} observes 77.91\% Top-1 accuracy on ImageNet when the patch size is 16$\times$16 while only 73.38\% is reached if the patch size is 32$\times$32.
%
%
%
Modern ViT structures simply accept a fixed patch size \emph{w.r.t.}\ all input images when training a ViT model.
It remains unexplored to train ViT models with a larger patch size for an economic computation while injecting information of a smaller patch size to retain the performance.

Luckily, images are often filled with redundant regions, such as backgrounds.
This property has inspired many researchers to go further, and drop the less informative tokens after forwarding the token sequence to the networks.
Tang \emph{et al}.~\citep{tang2021patch} introduced a top-down token pruning paradigm.
DynamicViT~\citep{rao2021dynamicvit} and IA-RED$^2$~\citep{pan2021ia} learn to score each token with a learnable prediction module, while EViT~\citep{liang2022evit} utilizes off-the-shelf class attention to measure token importance. In DVT~\citep{wang2021not}, multiple ViTs are cascaded and each image's preserved token number is decided by an early-existing policy.
Despite that these token drop methods decrease computation costs, they sacrifice recognition accuracy.
For example, the recent PS-ViT~\citep{tang2021patch} and Evo-ViT~\citep{liang2022evit} decrease the Top-1 performance of DeiT-S~\citep{touvron2021training} by 0.4\% although 1.6$\sim$2.0G FLOPs are saved on ImageNet.
Most existing methods train a ViT model on the premise of a fixed token keeping rate. It remains an open issue to train one ViT under multiple token keeping rates such that the case of more token drops for computation savings can benefit performance increase from that of fewer drops.

Above all, most existing methods are restricted to processing token sequence with a permanent patch size or excavating token redundancy with a fixed keeping rate. Once the training is finished, the inference process is deterministic, thus these methods result in a trained model with a static complexity, which not only bears a poor trade-off between performance and inference cost, but fails to support good hardware efficiency given that the hardware, even on the same workstation, is often equipped with different battery conditions or workloads at different periods.
%

We 
present a novel training paradigm that derives only one ViT model at a time but is endowed with a versatile ability of image recognition and its complexity can dynamically adapt to the current hardware resources. Figure\,\ref{framework} illustrates our training framework.
We make copies of an input image into multiple parallel branches each of which is split into local patches of a particular size. These patch branches are sequentially fed to a ViT model to take in multi-size patch information.
For each patch sequence, we also make efforts to excavate redundant regions from the perspective of training a network with multiple token keeping rates.
Consequently, to our best knowledge, this is the first study that can obtain one ViT model capable of recognizing images at multiple complexities in inference.
The trained ViT model in this paper, is termed as super vision transformer (SuperViT), where ``super'' refers to the ability to dispose of incoming patches of different sizes as well as preserving informative tokens with varying keeping rates.
When compared with the very recent token drop methods~\citep{rao2021dynamicvit,liang2022evit,pan2021ia,xu2022evovit}, our SuperViT has the following advantages:
(1) In contrast to these studies on token drops that sacrifice recognition accuracy, we observe that our training paradigm provides a better recognition ability. For example, the backbone network of DeiT-S increases by 0.2\% on ImageNet even when 50\% of tokens are removed.
(2) Our SuperViT provides better hardware efficiency since it allows a fast accuracy-efficiency trade-off by adapting the patch size of input images as well as token keeping rate to fitting the currently available hardware resources.

\section{Related Work}
The pioneering ViT work dates back to~\citep{dosovitskiy2020image} that applies a pure transformer in natural language processing~\citep{vaswani2017attention} to image classification.
However, due to the lack of inductive bias, its state-of-the-art performance depends too much on a very large-scale data corpus such as JFT-300M~\citep{sun2017revisiting}.
To overcome the necessity of large datasets for training the transformer, DeiT~\citep{touvron2021training} introduces token-based distillation and strong data augmentation.
Since then, based on the main spirit of ViTs, substantial breakthroughs have been made in computer vision society.
Except for the studies on token drops discussed in Sec.\,\ref{introduction}, we further briefly revisit some related studies below and also encourage the readers to go further on the survey paper~\citep{khan2021transformers,han2022survey} for a more comprehensive overview.

Albeit 
ViT's advantage in capturing long-range relations between image patches, it fails to model local patch information.
This motivates the researchers to compute self-attention within the local/windowed region.
Swin Transformer~\citep{liu2021swin} devises a shifted window partitioning to realize cross-window connections while the attention is performed upon each local window.
TNT~\citep{han2021transformer} leverages an inner block to strengthen the interactions of pixel information within each patch.
Twins~\citep{chu2021twins} alternates locally-grouped self-attention and global sub-sampled attention layer-by-layer.
These works not only bring back local information but also improve the efficiency of ViT models since the whole computation of global attention is avoidable.

Apart from 
the above studies, there are also many variants to follow the footprints of convolutional neural networks (CNNs).
Shuffle Transformer~\citep{huang2021shuffle} implements information flow among windows by a spatial shuffle operation as in ShuffleNet~\citep{zhang2018shufflenet}.
Alike to the depthwise convolution and pointwise convolution in MobileNet~\citep{howard2017mobilenets}, SepViT~\citep{li2022sepvit} devises depthwise self-attention to capture local representation within each window and pointwise self-attention to build connections among windows.
By incorporating the pyramid structure from CNNs, PVT~\citep{wang2021pyramid} serves as a versatile backbone for many dense prediction tasks.
DeformableViT~\citep{xia2022vision} equips with a deformable self-attention module in line with deformable convolution~\citep{dai2017deformable} to enable flexible spatial locations conditioned on input data.
ViT-Slim~\citep{chavan2022vision} searches for a sub-transformer network across three dimensions of input tokens, MHSA and MLP modules. 
%
%
%

Also, constructing the transformer with convolutions  can be a more straightforward way to solve the inductive bias.
CPVT~\citep{chu2021conditional} uses a convolution layer to replace the learnable positional embedding for fine-level feature encoding.
CMT~\citep{guo2022cmt} hybridizes CNNs and transformers to respectively capture local and global information, which promotes the ability of network representation.
Graham~\emph{et al}.~\citep{graham2021levit} revisited principles from extensive CNN studies. Consequently, LeViT, a transformer architecture inspired by convolutional approaches, is proposed with a trade-off between accuracy performance and inference speeds.
Mobile-Former~\citep{chen2021mobile} parallelizes MobileNets~\citep{howard2017mobilenets} and transformers with a two-way bridge to fuse local and global features.

\section{Methodology}

\subsection{Overview\label{overview}}
Vision transformer (ViT) breaks down an image $\mathit{I} \in \mathbb{R}^{H \times W \times C}$ into a set of $N$ local patches with shape of $P \times P \times C$. We have $N = \nicefrac{(H \cdot W)}{(P \cdot P)}$. These patches are linearly projected into $D$-dimensional token vectors. %
Also, an extra [class] token, which learns global image information and is responsible for the final classification, is added to the token sets. As results, the input token sequence of a ViT model can be represented as:
\begin{align}\label{embedding_equal}
\mathbf{X}^0 = [\mathbf{x}_{cls}^0;\mathbf{x}^0_1;...;\mathbf{x}^0_N] + \mathbf{E}_{pos} ,
\end{align}
where $\mathbf{x}^0_{cls} \in \mathbb{R}^{D}$ denotes [class] token and $\mathbf{x}^0_{i}$ represents the token of the $i$-th patch with $i > 0$. The $\mathbf{E}_{pos}$ denotes the learnable position embedding.
Then, all tokens are fed into a ViT model $\mathcal{T}$ with $L$ sequentially-stacked transformer encoders, each of which consists of a multi-head self-attention (MHSA) layer and a feed-forward network (FFN).
Denoting $\mathbf{X}^{l-1}$ and $\mathbf{X}^{l}$ as the input and output of the $l$-th transformer encoder, the processing of MHSA and FFN is formulated as:\footnote{Layer normalization is usually inserted before MHSA and FFN. We omit it here for brevity.}
\begin{align}
    &\mathbf{Y}^l = \mathbf{X}^{l-1} + \text{MHSA}(\mathbf{X}^{l-1}), \\
    &\mathbf{X}^l = \mathbf{Y}^l + \text{FFN}(\mathbf{Y}^l).
\end{align}

Usually, FFN consists of two fully-connected layers with a non-linear mapping inserted in-between such as GELU~\citep{hendrycks2016gaussian}.
In MHSA, the input tokens are linearly mapped to three matrices including a query $\mathbf{Q}$, a key $\mathbf{K}$ and a value $\mathbf{V}$, and the MHSA can be formulated as:
\begin{equation}
\begin{aligned}
&\text{MHSA}(\mathbf{X}^{l-1})
  \\&  =\text{Concat}\Big[ \text{Attention}(\mathbf{Q}^{l,h}, \mathbf{K}^{l,h}) \mathbf{V}^{l,h} \Big]_{h=1}^H \mathbf{W}^l,
\end{aligned}
\end{equation}
where 
$\text{Concat}[\cdot]$ concatenates its inputs and we have $\mathbf{Q}^l = \text{Concat}[\mathbf{Q}^{l,h}]_{h=1}^H$, $\mathbf{K}^l = \text{Concat}[\mathbf{K}^{l,h}]_{h=1}^H$, $\mathbf{V}^l = \text{Concat}[\mathbf{V}^{l,h}]_{h=1}^H$.
%
$\mathbf{W}^l$ is a projection matrix.
%
$\text{Attention}(\cdot , \cdot)$ is computed as:
\begin{equation}\label{attention}
\begin{aligned}
\text{Attention}(\mathbf{Q}^{l,h}, \mathbf{K}^{l,h}) &= [\mathbf{a}^{l,h}_{cls}; \mathbf{a}^{l,h}_1;...;\mathbf{a}^{l,h}_N] \\&= \text{Softmax} \big( \frac{\mathbf{Q}^{l,h}(\mathbf{K}^{l,h})^T}{\sqrt{D}} \big).
\end{aligned}
\end{equation}

In particular, $\mathbf{a}^l_{cls} = \sum_{h=1}^H \mathbf{a}_{cls}^{l,h}$ is known as the attention from [class] token to all patch tokens and often used to determine the information richness of each token~\citep{liang2022evit,xu2022evovit,chen2022coarse}. 
After a series of MHSA-FFN transformations, the [class] token $\mathbf{x}_L^{cls}$ is extracted from the $L$-th transformer encoder and utilized for object category prediction. Taking classification as an example, $\mathbf{x}_L^{cls}$ is fed to a classifier consisting of a fully-connected layer and a softmax layer.
Therefore, given an $L$-layer ViT model $\mathcal{T}$ with the input token sequence $\mathbf{X}_0 \in \mathbb{R}^{N \times D}$, the category prediction distribution is obtained as:
\begin{align}
    \mathbf{p} = \mathcal{T}(\mathbf{X}^0) = \text{Softmax}\big(\text{FC}(\mathbf{x}^L_{cls})\big).
\end{align}

\textbf{Discussion}. The ViT model benefits from the MHSA that models the long-range dependencies between the input tokens. However, the main computational complexity also stems from the MHSA layer~\citep{dosovitskiy2020image, liu2021swin} as $\mathcal{O}(\text{MHSA}) = 4ND^2 + 2N^2D.
$
We can see that the complexity of MHSA increases quadratically 
in 
the number of incoming tokens $N$. Consequently, the MHSA has become the computation bottleneck in the ViT model.
The first naive manner is to reduce the transformer's input sequence length by enlarging the size of split patches. However, as analyzed in Sec.\,\ref{introduction}, the performance of a ViT model is closely correlated with the patch size as well. 
The second intuitive way is to discard tokens considered less informative in network forward, which is observed to deteriorate the accuracy as discussed in Sec.\,\ref{introduction}.

We analyze that existing methods~\citep{rao2021dynamicvit,liang2022evit,jiang2021all,touvron2021training} are overfitting to a static complexity since they process token sequence with a permanent length or excavating token redundancy with a fixed keeping rate. The resulting ViT models suffer a poor accuracy-speed trade-off, as well as fail to support good hardware efficiency.
This impels us to learn one versatile transformer of recognizing images at multiple complexities, named SuperViT, details of which are given below.

\subsection{Super Vision Transformer}

We aim to train only one ViT model
that maintains computational costs at different levels.
We present how our SuperViT is empowered with a better capability of image recognition and can adapt to the current availability of hardware resources.
The overview of SuperViT is depicted in Figure\,\ref{framework}, which mainly includes a part of multi-size patch splitting and a part of multi-token keeping rate.
%

\subsubsection{Multi-Size Patch Splitting\label{multi-size-patch}}

The contradiction exists that larger-size patches reduces the computational costs while the recognition benefits from smaller-size patches.
To maintain good performance at low computation, we propose to inject information of larger-size patches into the training of a ViT model with a smaller-size input.
That is, we intend to equip SuperViT with the ability to solve incoming patches of multiple sizes.

%
To that end, 
as shown in Figure\,\ref{framework}, the input image is copied into $G$ parallel branches and each branch is responsible for a particular patch size in the image splitting. Consequently, we have a patch size set $\{ P_g \times P_g \}_{g=1}^G$ where $P_g > P_{g+1}$.
Then,  an input image $I \in \mathbb{R}^{H \times W \times C}$ is split into $G$ patch sets and the $g$-th set consists of local patches with 
a 
shape of $P_g \times P_g \times C$, leading to a sequence length of $N_g = \frac{H \cdot W}{P_g \cdot P_g}$.
Modern ViT structures simply accept a
fixed 
size of input patches \emph{w.r.t.} all training images such that the patch sequence can be embedded into tokens of the same dimension for training a ViT model in parallel.
To embed patches of different shapes into the token vectors in a $D$-dimensional space, a simple 
approach 
is to consider individual patch embedding layers for each patch set~\citep{zhu2021make, wang2021not}, which however, increases the parameters.
Instead, we choose the economical bilinear interpolation to downsample/upsample these local patches for a shape alignment first. Then, these aligned patches are fed to a shared embedding layer to obtain the input token sequence $\mathbf{X}_g^0$ for the $g$-th patch set where $\mathbf{X}^0_g$ is defined as:
\begin{equation}
    \mathbf{X}_g^0 = \big[ \mathbf{x}_{g,cls}^0; \mathbf{x}_{g,1}^0;...; \mathbf{x}_{g,N_g}^0 \big] + \mathbf{E}_{g, pos},
\end{equation}
where $\mathbf{x}^0_{g, cls}$ and $\mathbf{x}^0_{g, i} (i > 0)$ represent the class token and the $i$-th token in the $g$-th input patch set.
Then, we feed the token sequence $\{\mathbf{X}_g^0\}_{g=1}^G$ to the ViT one-by-one for a series of MHSA-FFN transformations described in Sec.\,\ref{overview}. Finally, we obtain a distribution set of category predictions $\{ \mathbf{p}_g \}_{g=1}^G$, where each prediction $\mathbf{p}_g$ is derived by:
\begin{align}
    \mathbf{p}_g = \mathcal{T}(\mathbf{X}_g^0) = \text{Softmax}\big(\text{FC}(\mathbf{x}_{g,cls}^L)\big).
\end{align}

Then, the prediction distributions $\{ \mathbf{p}_g \}_{g=1}^G$ can be used to formulate the learning objective, such as cross-entropy loss with the ground-truth labels for ViT training.
The trade-off between computational budget and accuracy performance can hardly be made if a ViT model is trained under a single patch size such as $P_1 \times P_1$.
Luckily, our SuperViT can well enhance the performance of $P_1 \times P_1$ at the test stage since information of smaller-size patches is injected during network training. 
Also, the performance of smaller-size patches can be enhanced by larger-size ones since images of different complexities require different patch sizes to be correctly classified~\citep{wang2021not,chen2022coarse}.

%
%
%
%
%

\subsubsection{Multi-Token Keeping Rate}
We further 
spend 
efforts to reduce computational costs of our SuperViT given that redundant regions widely exist in image content.
As discussed in Sec.\,\ref{introduction}, most existing studies on dropping less informative tokens suffer poor performance since they train the ViT model with a fixed token keeping rate, which fails to adapt to images of different complexities.
Therefore, we propose to strengthen our SuperViT with the ability to preserve informative tokens with multiple keeping rates.

As shown in Figure\,\ref{framework}, we predesignate a set of token keeping rates $\{ \eta_m \}_{m=1}^M$ where $\eta_m > \eta_{m+1}$, and $0 < \eta_m < 1$ for $m > 1$ which means the top-($\eta_m\cdot N$) informative tokens from the input sequence will be preserved. We define $\eta_1 = 1$, which indicates the case of preserving all tokens is always performed.
Then, for token sequence in the $l$-th layer $\mathbf{X}_g^{l} = [\mathbf{x}^{l}_{g, cls}; \mathbf{x}^{l}_{g, 1};...;\mathbf{x}^{l}_{g, N_g}]$, we determine the information richness of each token using $\mathbf{a}^l_{cls}$.

As defined in Eq.~(\ref{attention}), the $i$-th entry of $\mathbf{a}^l_{cls}$ determines how much information of the $i$-th token $\mathbf{x}^l_{g, i}$ is fused into the class token $\mathbf{x}^{l}_{g, cls}$~\citep{liang2022evit}. 
It thus has become an indicator to reflect the information richness of each token in many existing studies~\citep{liang2022evit,xu2022evovit,chen2022coarse}.
We focus more on training one versatile ViT model proficient in processing multi-token keeping rates, thus following existing studies, we directly preserve tokens with larger attention values in this paper.

With the predefined keeping rate set $\{ \eta_m \}_{m=1}^M$, we sequentially feed the incoming token sequence for MHSA-FFN transformations. The $m$-th forward propagation is constrained by token keeping rate $\eta_m$ and the resulting category prediction distribution is formulated as:
\begin{align}
    \mathbf{p}_{g,m} = \mathcal{T}(\mathbf{X}^0_{g} \mid \eta_m) = \text{Softmax}\big(\text{FC}(\mathbf{x}_{g, cls}^L \mid \eta_m)\big),
\end{align}
%


For $\eta_1 = 1$, we have $\mathbf{p}_{g, 1} = \mathbf{p}_{g} = \mathcal{T}(\mathbf{X}_g^0)$.
Similar to multi-size patch splitting, training a ViT model with multiple token keeping rates also benefits the performance while the less informative tokens are removed for computation savings.

\subsection{Training Objective}\label{training_objective}
Based on the proposed multi-size patch splitting and multi-token keeping rate, our SuperViT would result in a total of
$G \times M$
different computational costs.  
Inspired by the one-shot training settings in traditional CNNs~\citep{yu2018slimmable, cai2019once, yang2020mutualnet}, in each training iteration, we first freeze updating our SuperViT and sequentially forward a token sequence with a particular patch size as well as token keeping rate to derive the corresponding category prediction distribution set $\{ \mathbf{p}_{g,m} \}_{g=1:G, m=1:M}$.
Among these predictions, it is expected that $\mathbf{p}_{g,1}$ can fit best with the ground-truth label $\mathbf{y}$ since no token drop is performed in this case. Thus, we propose to supervise the learning of SuperViT with the ground-truth labels in the case of no performing token drop, and use $\mathbf{p}_{g,1}$ as a knowledge hint to guide the learning of SuperViT in the case of performing token drops, leading to our training objective as:
%
%
\begin{equation}\label{objective}
    \mathcal{L} = \sum_{g = 1}^G \text{CE}(\mathbf{p}_{g, 1}, \mathbf{y}) + \sum_{g=1}^G\sum_{m=2}^M \text{KL}(\mathbf{p}_{g, m}, \mathbf{p}_{g, 1}),
\end{equation}
where $\text{CE}(\cdot,\cdot)$ denotes the cross-entropy loss and $\text{KL}(\cdot,\cdot)$ represents the Kullback-Leibler divergence.
After calculating the training loss, we switch on the gradient computing to update our SuperViT.
However, looping over all the $G \cdot M$ cases causes heavy training burden.
Instead, we offer an alternative where only four complexities are considered in each iteration to reduce the training consumption. 

\textbf{Hardware Efficiency}.
Our training paradigm results in one single ViT model executable to recognize images at different computational costs. It can be well deployed upon various hardware platforms with different resource constraints. Even for a given hardware device, our SuperViT permits instant and adaptive accuracy-efficiency trade-offs at runtime once the battery conditions or workloads change by simply modifying the image patch size and token keeping rate. 
Therefore, the hardware efficiency of our SuperViT is very advantageous over the traditional scenario that has to train numerous ViT models in advance, and dynamically download an appropriate model and offload the existing one.

\begin{table*}[!t]
\begin{center}
\caption{\label{throughput_table} Efficacy comparison between our SuperViT and its backbones including DeiT-T, DeiT-S, LV-ViT-T and LV-ViT-S. The proposed SuperViT is a framework on top of these backbones. Experiments are conducted on ImageNet.}
\resizebox{\textwidth}{!}{
\begin{tabular}{cccccc}
\hline\noalign{\smallskip}
\multirow{2}{*}{Model} & {Sequence} & {Keeping} &  Top-1 & FLOPs{\color{blue}$\downarrow$} & Throughput{\color{blue}$\uparrow$}\\
 &Length &Rate & Acc.{\color{blue}$\uparrow$} (\%) & (G) &(img./s) \\
\hline
DeiT-S (Backbone)  & 14$\times$14 & 1.0 & 79.8 & 4.6 & 2461 \\
SuperViT & 12$\times$12 & 0.7 &  79.6{\color{blue}($-0.2$)} & 2.2{\color{blue}($\downarrow$52\%)} & 4996{\color{blue}($\uparrow$2.03$\times$)} \\
SuperViT & 12$\times$12 & 1.0 &  79.9{\color{blue}($+0.1$)} & 3.3{\color{blue}($\downarrow$28\%)} &  3371{\color{blue}($\uparrow$1.37$\times$)} \\
SuperViT & 14$\times$14 & 0.5 &  80.0{\color{blue}($+0.2$)} & 2.3{\color{blue}($\downarrow$50\%)} & 4767{\color{blue}($\uparrow$1.94$\times$)} \\
SuperViT & 14$\times$14 & 0.7 &  80.5{\color{blue}($+0.7$)} & 3.0{\color{blue}($\downarrow$35\%)} & 3654{\color{blue}($\uparrow$1.48$\times$)} \\
SuperViT & 14$\times$14 & 1.0 &  80.6{\color{blue}($+0.8$)} & 4.6{\color{blue}($\downarrow$0\%)} & 2461{\color{blue}($\uparrow$1.00$\times$)} \\
\hdashline
DeiT-T & 14$\times$14 & 1.0 & 72.2 & 1.3 & 5013 \\
SuperViT & 8$\times$8 & 0.5 & 73.9{\color{blue}($+1.7$)}  & 0.7{\color{blue}($\downarrow$46\%)}  &  13548{\color{blue}($\uparrow$2.70$\times$)}  \\
SuperViT & 8$\times$8 & 0.7 & 75.3{\color{blue}($+3.1$)}  & 1.0{\color{blue}($\downarrow$23\%)} &  10669{\color{blue}($\uparrow$2.13$\times$)} \\
SuperViT & 8$\times$8 & 1.0 & 75.8{\color{blue}($+3.6$)} & 1.4{\color{blue}($\uparrow$8\%)} &  7727{\color{blue}($\uparrow$1.54$\times$)} \\
SuperViT & 10$\times$10 & 0.5 & 77.3{\color{blue}($+5.1$)}  & 1.2{\color{blue}($\downarrow$8\%)} & 9657{\color{blue}($\uparrow$1.93$\times$)} \\
SuperViT & 10$\times$10 & 0.7 & 78.3{\color{blue}($+6.1$)}  & 1.5{\color{blue}($\uparrow$15\%)} & 7567{\color{blue}($\uparrow$1.51$\times$)} \\
SuperViT & 10$\times$10 & 1.0 & 78.5{\color{blue}($+6.2$)}  & 2.3{\color{blue}($\uparrow$77\%)} &  5173{\color{blue}($\uparrow$1.03$\times$)} \\
SuperViT & 12$\times$12 & 0.5 & 78.9{\color{blue}($+6.7$)}  & 1.7{\color{blue}($\uparrow$31\%)} &  6308{\color{blue}($\uparrow$1.26$\times$)} \\\hline
DeiT-T (Backbone)   & 14$\times$14 & 1.0 & 72.2 &1.3 &5013 \\
SuperViT   &14$\times$14 &1.0 &73.1{\color{blue}(+0.9)} &1.3{\color{blue}($\downarrow$0\%)} &5013($\uparrow$1.00$\times$) \\
\hline
LV-ViT-S (Backbone) & 14$\times$14 & 1.0 & 83.3 & 6.6 & 1748 \\
SuperViT & 12$\times$12 & 0.7 & 82.6{\color{blue}($-0.7$)}  & 3.2{\color{blue}($\downarrow$52\%)} & 3565{\color{blue}($\uparrow$2.04$\times$)} \\
SuperViT & 12$\times$12 & 1.0 & 82.9{\color{blue}($-0.4$)}  & 4.7{\color{blue}($\downarrow$29\%)} & 2357{\color{blue}($\uparrow$1.35$\times$)} \\
SuperViT & 14$\times$14 & 0.7 &  83.2{\color{blue}($-0.1$)} & 4.3{\color{blue}($\downarrow$35\%)} & 2684{\color{blue}($\uparrow$1.54$\times$)} \\
SuperViT & 14$\times$14 & 1.0 &  83.5{\color{blue}($+0.2$)} & 6.6{\color{blue}($\downarrow$0\%)} & 1748{\color{blue}($\uparrow$1.00$\times$)} \\
\hdashline
LV-ViT-T & 14$\times$14 & 1.0 & 79.1 & 2.9 & 3178 \\
SuperViT & 8$\times$8 & 0.5 & 76.6{\color{blue}($-2.5$)}  & 1.1{\color{blue}($\downarrow$62\%)} & 9968{\color{blue}($\uparrow$3.14$\times$)} \\
SuperViT & 8$\times$8 & 0.7 & 79.8{\color{blue}($+0.7$)}  & 1.4{\color{blue}($\downarrow$52\%)} & 7836{\color{blue}($\uparrow$2.47$\times$)} \\
SuperViT & 8$\times$8 & 1.0 & 80.7{\color{blue}($+1.6$)}  & 2.0{\color{blue}($\downarrow$31\%)} & 5487{\color{blue}($\uparrow$1.72$\times$)} \\
SuperViT & 10$\times$10 & 0.5 & 79.8{\color{blue}($+0.7$)} & 1.7{\color{blue}($\downarrow$41\%)} & 6792{\color{blue}($\uparrow$2.13$\times$)} \\
SuperViT & 10$\times$10 & 0.7 & 81.7{\color{blue}($+2.6$)}  & 2.2{\color{blue}($\downarrow$24\%)} & 5302{\color{blue}($\uparrow$1.67$\times$)} \\
SuperViT & 10$\times$10 & 1.0 & 82.2{\color{blue}($+3.1$)}  & 3.3{\color{blue}($\uparrow$14\%)} & 3615{\color{blue}($\uparrow$1.14$\times$)} \\
SuperViT & 12$\times$12 & 0.5 & 81.1{\color{blue}($+2.0$)}  & 2.5{\color{blue}($\downarrow$14\%)} & 4524{\color{blue}($\uparrow$1.42$\times$)} \\
SuperViT & 14$\times$14 & 0.5 &  82.1{\color{blue}($+3.0$)} & 3.4{\color{blue}($\uparrow$17\%)} & 3368{\color{blue}($\uparrow$1.06$\times$)} \\
\hline
LV-ViT-T (Backbone) &14$\times$14 &1.0 &79.1 &2.9 &3178 \\
SuperViT &14$\times$14 &1.0  &79.7{\color{blue}(+0.6)}  &2.9{\color{blue}($\downarrow$0\%)} &3178{\color{blue}($\uparrow$1.00$\times$)} \\
\hline
\end{tabular}
}
\end{center}
\end{table*}

Though its three extra forward passes seem to increase training cost, SuperViT results in multiple subnets of good performance. Thus, more fair comparison should be executed by looking at the total costs for the same number of individually trained networks, which is provided in Sec.\,\ref{ablation} and demonstrates not only better performance of our subnets, but also more economic training consumption.
%

\section{Experiments\label{experiments}}

\subsection{Implementation Details}\label{details}
We evaluate our SuperViT and compare it against 
state-of-the-art 
on ImageNet~\citep{deng2009imagenet}. 
Following existing studies on ViT compression~\citep{rao2021dynamicvit,xu2022evovit,liang2022evit}, we use DeiT (w/o distillation)~\citep{touvron2021training} and LV-ViT~\citep{jiang2021all} as the backbones. All the training strategies, such as data augmentation, regularization and optimizer, strictly follow the original settings of DeiT~\citep{touvron2021training} and LV-ViT~\citep{jiang2021all}. 
The sequence length in our multi-size splitting includes \{8$\times$8, 10$\times$10, 12$\times$12, 14$\times$14\} and the multi-token keeping rate includes \{1.0, 0.7, 0.5\}.
We remove less informative tokens at the 4-th,7-th,10-th blocks for both DeiT and LV-ViT to follow compared methods, and train SuperViT on a workstation with 4 NVIDIA A100 GPUs.

\subsection{Performance Results}

\subsubsection{Comparison with Backbones}
To show the efficacy of our training paradigm, we first list our SuperViT under different sequence lengths (patch sizes) and token keeping rates in Table\,\ref{throughput_table} and compare it with its backbones of DeiT-S and LV-ViT-S. For fair comparison, we also compare with DeiT-T and LV-ViT-T backbones since our computational costs are tiny in the cases of very small sequence lengths and keeping rates.
Our efficacy is measured from two aspects including the Top-1 accuracy to reflect its effectiveness, and FLOPs consumption and model throughput to reflect its efficiency.
The model throughput shows the number of processed images per second on a single A100 GPU~\citep{wang2021not,liang2022evit}.
For an accurate throughput estimation, we repeatedly feed each model with a batch size of 512 for 50 times. Then, the practical throughput is computed as $\nicefrac{512 \times 50}{\text{total inference time}}$.

From Table\,\ref{throughput_table}, we observe 
significant 
accuracy increase when maintaining similar FLOPs consumption and throughput with backbones.
For example, with FLOPs of 4.6G and throughput of 2461, our SuperViT (14$\times$14, 1.0) increases the accuracy of DeiT-S backbone from 79.8\% (2-nd row of Table\,\ref{throughput_table}) to 80.6\% (7-th row of Table\,\ref{throughput_table}), leading to 0.8\% improvement. 
Also, from the 22-nd row of Table\,\ref{throughput_table}, our SuperViT (14$\times$14, 1.0) gains additional 0.2\% accuracy improvement when using LV-ViT-S as the backbone.
Besides, when maintaining similar accuracy, SuperViT merits in its significant FLOPs reduction (throughput increase).
For example, with accuracy of 79.9\% (SuperViT, 4-th row of Table\,\ref{throughput_table}) and 79.8\% (DeiT-S, 2-nd row of Table\,\ref{throughput_table}), SuperViT saves 28\% FLOPs and obtains 1.37$\times$ throughput increase, meanwhile SuperViT reduces 35\% FLOPs and increases throughput by 1.54$\times$ with accuracy of 83.2\% (21-st row of Table\,\ref{throughput_table}) over 83.3\% of LV-ViT-S (18-th row of Table\,\ref{throughput_table}).
Compared to the tiny version of DeiT and LV-ViT, our SuperViT not only reduces FLOPs and increases throughput, but also significantly enhances the performance.

\begin{figure}[!t]
\begin{center}
\includegraphics[width=\linewidth]{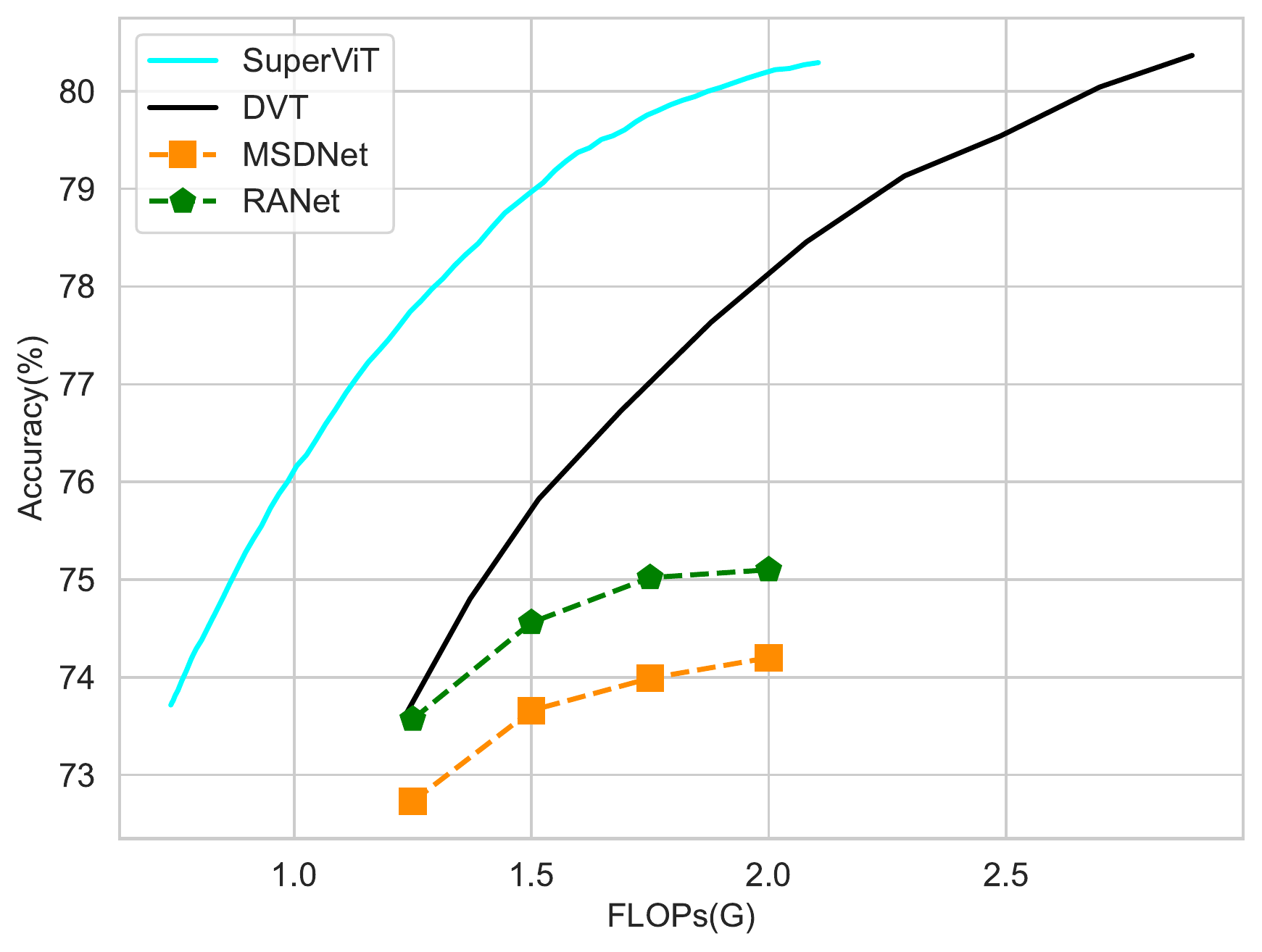}
\end{center}
\caption{\label{early_exit_compare_fig}Comparison between SuperViT and early-exiting compression methods including MSDNet, RANet and DVT. The proposed SuperViT uses DeiT-S as the backbone. Experiments are conducted on ImageNet.}
\end{figure}

It is worth noting that, for LV-ViT-S, SuperViT increases the accuracy only by 0.2\%, comparing to the significant improvement of 0.9\% for DeiT-T, 0.8\% for DeiT-S and 0.6\% for LV-ViT-T. This is due to the fact that LV-ViT-S is a relatively larger network that already achieves a very good performance of 83.3\%, upon which, it is challenging to gain much better performance. Given this, the 0.2\% improvement is also meaningful.

\subsubsection{Comparison with Compressed Models}
We continue to compare our efficacy with many studies on model compression including early-existing compression~\citep{huang2017multi,yang2020resolution,wang2021not} and token drop compression~\citep{pan2021ia,yin2022avit,chavan2022vision,liang2022evit,xu2022evovit,rao2021dynamicvit,tang2021patch}.
Early-existing compression methods dynamically choose different inference paths based on a particular criterion.
Our SuperViT can also implement early exist since it adapts to multiple complexities. 
In this part, we give a simple scenario where the cheapest computation (8$\times$8, 0.5) is always used to predict an incoming image and a second one (14$\times$14, 0.7) is utilized again if the confidence of the first prediction is smaller than a threshold. By adjusting the threshold, our SuperViT achieves different accuracy-FLOPs trade-offs.
%
%
Figure\,\ref{early_exit_compare_fig} plots the performance of our SuperViT built upon DeiT-S and methods including CNN-based MSDNet~\citep{huang2017multi} and RANet~\citep{yang2020resolution}, as well as transformer-based DVT~\citep{wang2021not}. 
Our SuperViT consistently performs better than the state-of-the-art competitor DVT~\citep{wang2021not}. With similar accuracy, SuperViT results in significantly smaller FLOPs. This is attributed to the fact that the cheapest version of SuperViT already reaches good accuracy of 73.9\% in Table\,\ref{throughput_table}. Therefore, most images can be well recognized even at very small costs.

\begin{table}[!t]
\begin{center}
\caption{\label{compare_sota} Efficacy comparison between our SuperViT and off-the-shelf studies on token drops. For fair comparison, all compared methods use the same backbones including DeiT-S and LV-ViT-S with our proposed SuperViT. Experiments are conducted on ImageNet.}
\setlength{\tabcolsep}{3.0pt}
\begin{tabular}{cccc}
\hline\noalign{\smallskip}
\multirow{2}{*}{Methods} 
& \multirow{2}{*}{Pre-trained} & Top-1 & FLOPs \\
& &Acc. (\%) & (G) \\
\hline
DeiT-S (Backbone) & - &  79.8 & 4.6 \\
\textbf{SuperViT (Ours)} & \XSolidBrush & \textbf{80.6} & \textbf{4.6} \\
IA-RED$^2$ & \Checkmark & 79.1 & 3.2 \\
A-ViT & \Checkmark & 78.6 & 3.6 \\
ViT-Slim & \Checkmark & 79.9 & 3.1 \\
Evo-ViT & \XSolidBrush & 79.4 & 3.0 \\
EViT & \XSolidBrush &79.5 & 3.0 \\
\textbf{SuperViT (Ours)} & \XSolidBrush &  \textbf{80.5} & \textbf{3.0} \\
DynamicViT & \Checkmark & 79.3 & 2.9 \\
PS-ViT & \Checkmark & 79.4 & 2.6 \\
\textbf{SuperViT(Ours)} & \XSolidBrush & \textbf{80.0} & \textbf{2.3} \\
\hline
LV-ViT-S (Backbone) & - & 83.3 & 6.6 \\
\textbf{SuperViT (Ours)} & \XSolidBrush & \textbf{83.5} & \textbf{6.6} \\
DynamicViT & \Checkmark &  83.0 & 4.6 \\
EViT & \XSolidBrush & 83.0 & 4.7 \\
\textbf{SuperViT (Ours)} & \XSolidBrush & \textbf{83.2} & \textbf{4.3} \\
\hline
\end{tabular}
\end{center}
\end{table}

We go on the comparison with token drop compression, which reduces the number of tokens during network forwarding.
For fair comparison, we pick up the results of our SuperViT from Table\,\ref{throughput_table} and compare with the recent state-of-the-arts~\citep{pan2021ia,yin2022avit,chavan2022vision,liang2022evit,xu2022evovit,rao2021dynamicvit,tang2021patch}.
Results in Table\,\ref{compare_sota} manifest that our SuperViT well outperforms previous methods in both accuracy performance and FLOPs reduction by margins when DeiT-S and LV-ViT-S are used as backbones.
For example, upon DeiT-S, our SuperViT significantly outperforms the recent state-of-the-art EViT~\citep{liang2022evit} by 1.1\% when consuming the same FLOPs of 3.0G.
It is also worth stressing that, existing methods, some of which even heavily rely on a pre-trained model, deteriorate the baseline performance when performing token drops while our SuperViT leads to performance increase in most cases.
For example, we reduce 2$\times$ FLOPs of DeiT-S while increasing the Top-1 accuracy by 0.2\% and 0.7\% for 1.5$\times$ reduction.
These results well demonstrate the efficacy of our training paradigm to obtain only one ViT model empowered with the ability to recognize images at different levels of computational costs.

\begin{figure*}[!t]
\begin{center}
\includegraphics[width=\linewidth]{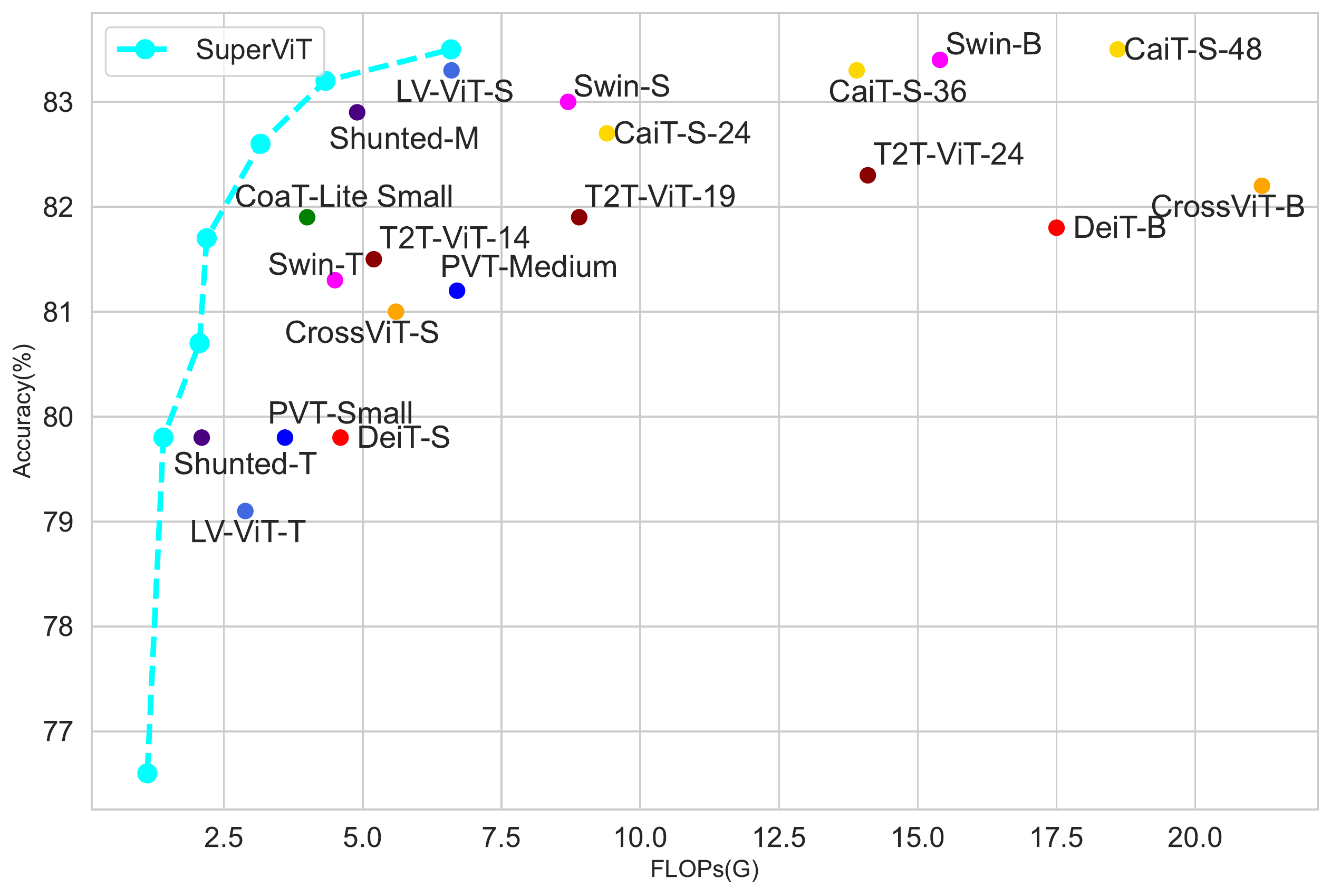}
\end{center}
\caption{\label{sota_compare_fig} Accuracy and FLOPs trade-off comparison between our SuperViT and popular ViT models. The proposed SuperViT uses LV-ViT-S as the backbone. Experiments are conducted on ImageNet.}
\end{figure*}

\begin{figure*}[!t]
  \centering
  \begin{subfigure}{0.32\linewidth}
    \includegraphics[width=\linewidth, height=0.6\linewidth]{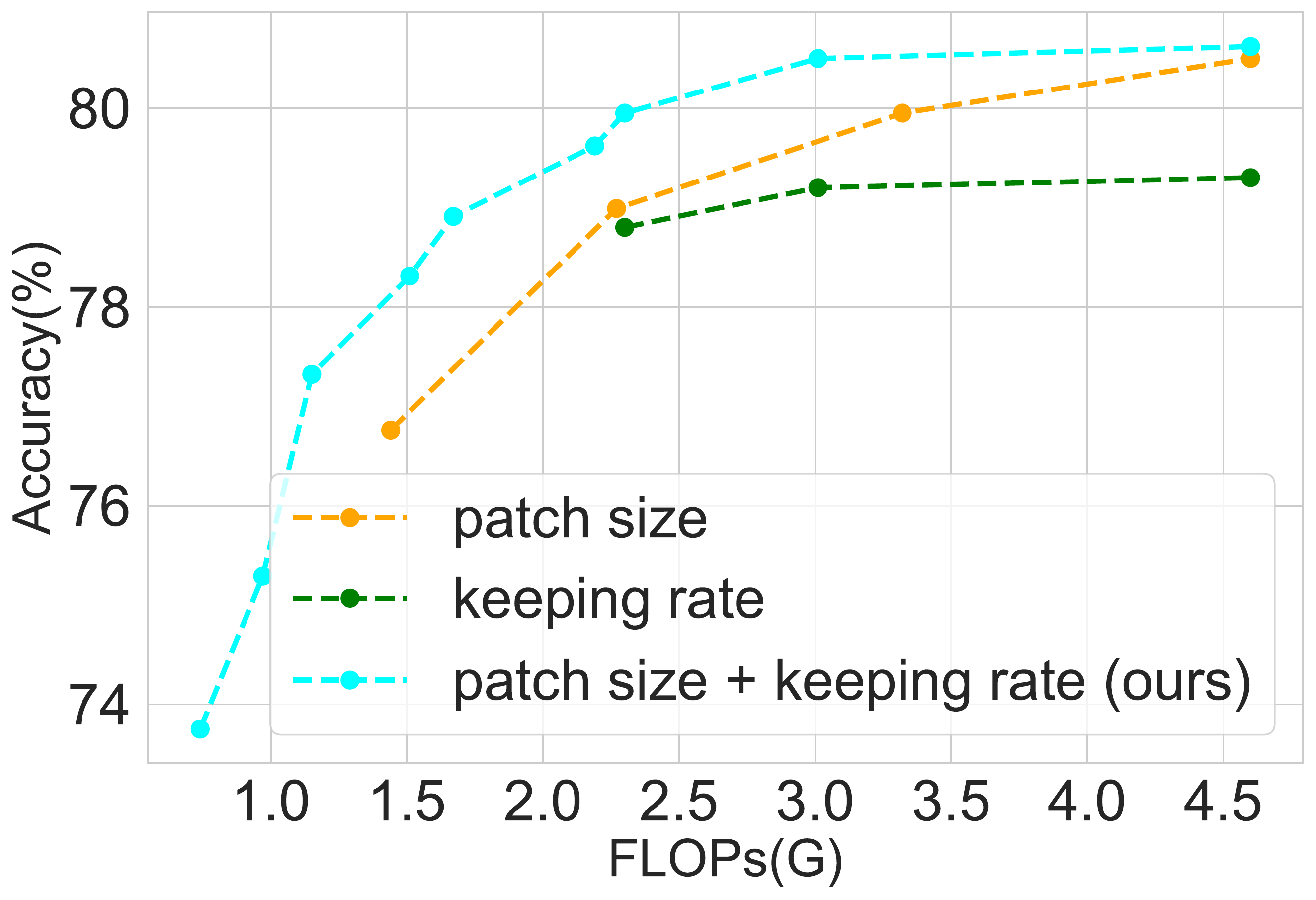}
    \caption{}
    \label{size_rate}
  \end{subfigure}
  \begin{subfigure}{0.32\linewidth}
    \includegraphics[width=\linewidth, height=0.6\linewidth]{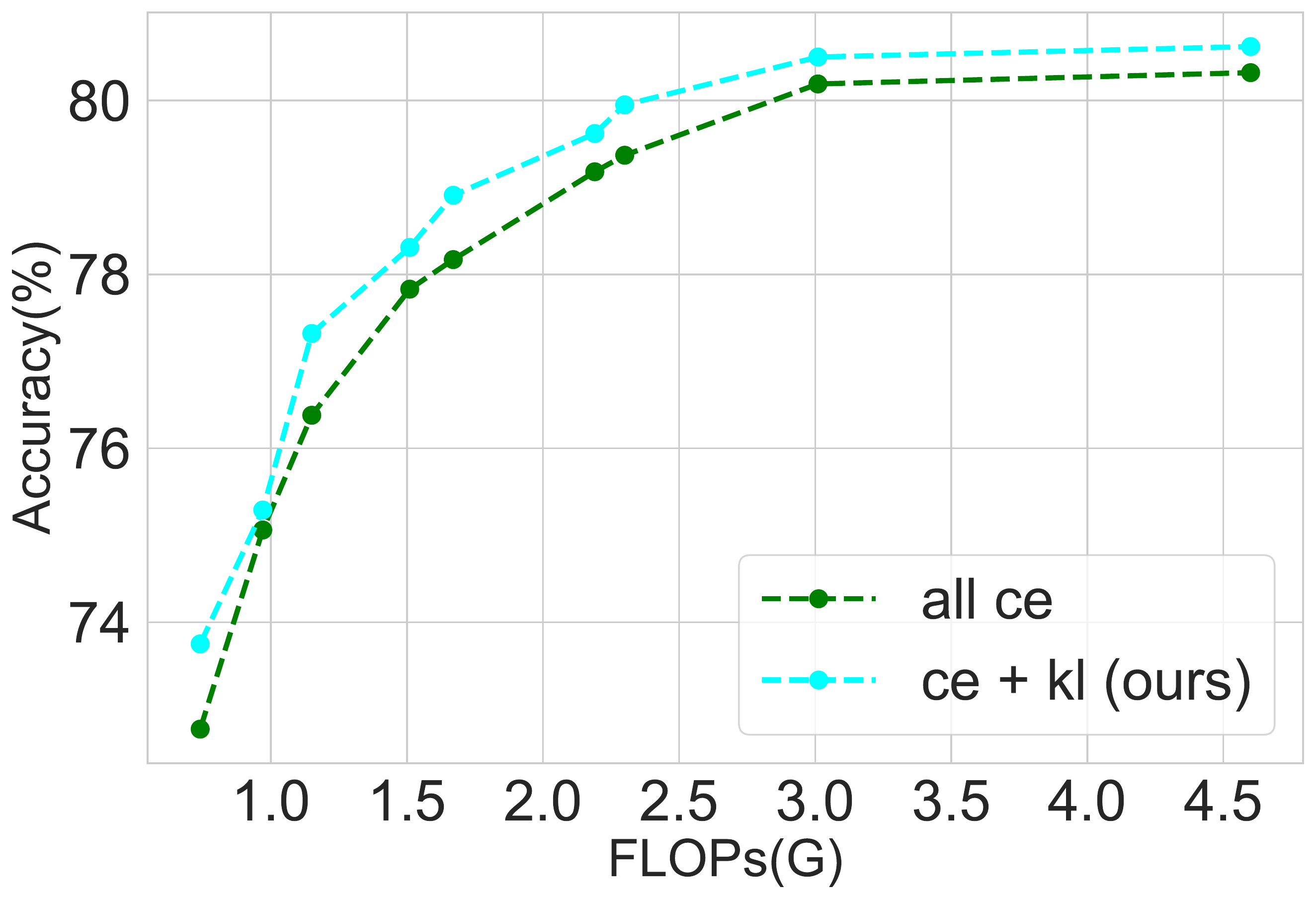}
    \caption{}
    \label{ce_kl}
  \end{subfigure}
  \begin{subfigure}{0.32\linewidth}
    \includegraphics[width=\linewidth, height=0.6\linewidth]{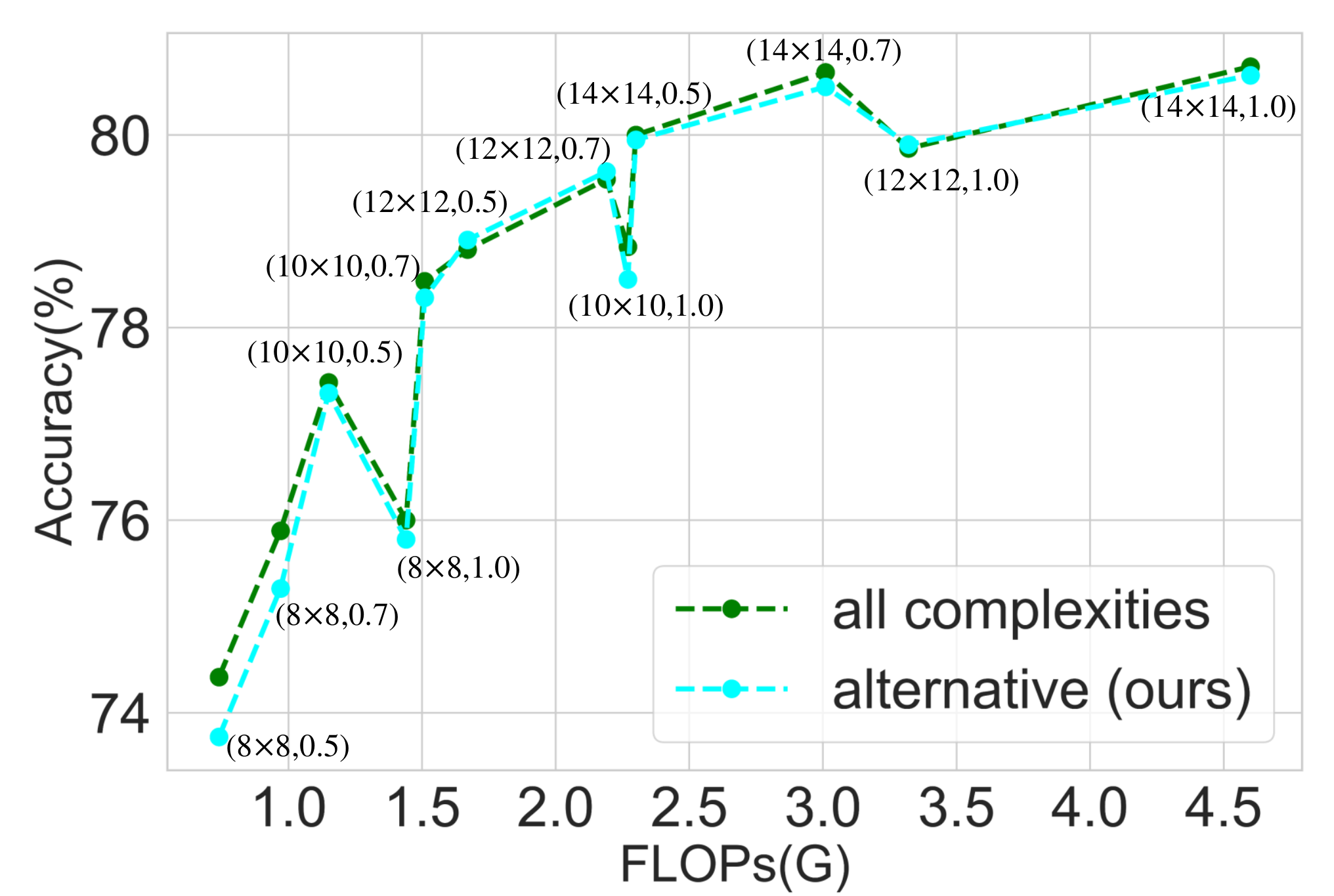}
    \caption{}
    \label{all_reduce}
  \end{subfigure}%
   \caption{Ablation studies of (a) multi-size and multi-token, (b) training objective, and (c) training complexities. We conduct ablations using DeiT-S as the backbone, and all the experiments are conducted on ImageNet.}
  \label{phenomenon}
\end{figure*}

\subsubsection{Comparison with ViT Models}
In Figure\,\ref{sota_compare_fig}, we compare the accuracy and FLOPs trade-off between our SuperViT built upon LV-ViT-S and various ViT models including DeiT~\citep{touvron2021training}, PVT~\citep{wang2021pyramid}, CoaT~\citep{xu2021co}, CrossViT~\citep{chen2021crossvit}, Swin~\citep{liu2021swin}, T2T-ViT~\citep{yuan2021tokens},
CaiT~\citep{touvron2021going} and
Shunted-ViT~\citep{ren2021shunted}. 
The accuracy-FLOPs trade-off of our SuperViT consists of models with different sequence lengths and keeping rates in Table\,\ref{throughput_table}.
Results in Figure\,\ref{sota_compare_fig} indicate that our SuperViT provides a better accuracy-FLOPs trade-off than majorities of existing ViT models. 
In fact, these ViTs focus on improving the structure of vanilla ViT or token interactions for accuracy improvement while our SuperViT introduces one new training paradigm. Thus, our SuperViT is orthogonal to these ViT variants, which increases the possibility of integrating our training paradigm into these ViT models.
For example, when constructing our SuperViT upon LV-ViT-S, a better accuracy-FLOPs trade-off can be observed in Figure\,\ref{sota_compare_fig}.


\subsection{Ablation Study\label{ablation}}
\textbf{Multi-size and Multi-token}.
In Figure\,\ref{size_rate}, we first analyze the effectiveness of multi-size patch splitting and multi-token keeping rate respectively. As can be seen from the figure, the combination of multi-size patch splitting and multi-token keeping rate results in consistent performance increase compared to these only considering one of them.
Our training objective in Eq.\,(\ref{objective}) adopts cross-entropy loss for the forwarding that does not perform token drops, while Kullback-Leibler divergence is used for the cases where token drops are performed. Figure\,\ref{ce_kl} compares our training objective with the situation where the cross-entropy is considered for all cases. Results show that our training objective is of more benefit to SuperViT.
Figure\,\ref{all_reduce} compares the performance of considering all $G \cdot M$ complexities and the alternative of using only four cases (see Sec.\ref{training_objective}). Results manifest that the former manifests slightly better performance.
Nevertheless, considering all complexities takes about 8.8 (DeiT-S backbone) and 14.4 (LV-ViT-S backbone) days to train SuperViT on four NVIDIA A100 GPUs, while they drastically decrease to 4.1 and 6.5 days by using only four subnets at each iteration of the forward propagation. Thus, we encourage to utilize the alternative if the training resources are not enough, which is also our standard to implement all experiments of this paper.

\begin{figure}[!t]
\begin{center}
\includegraphics[width=\linewidth]{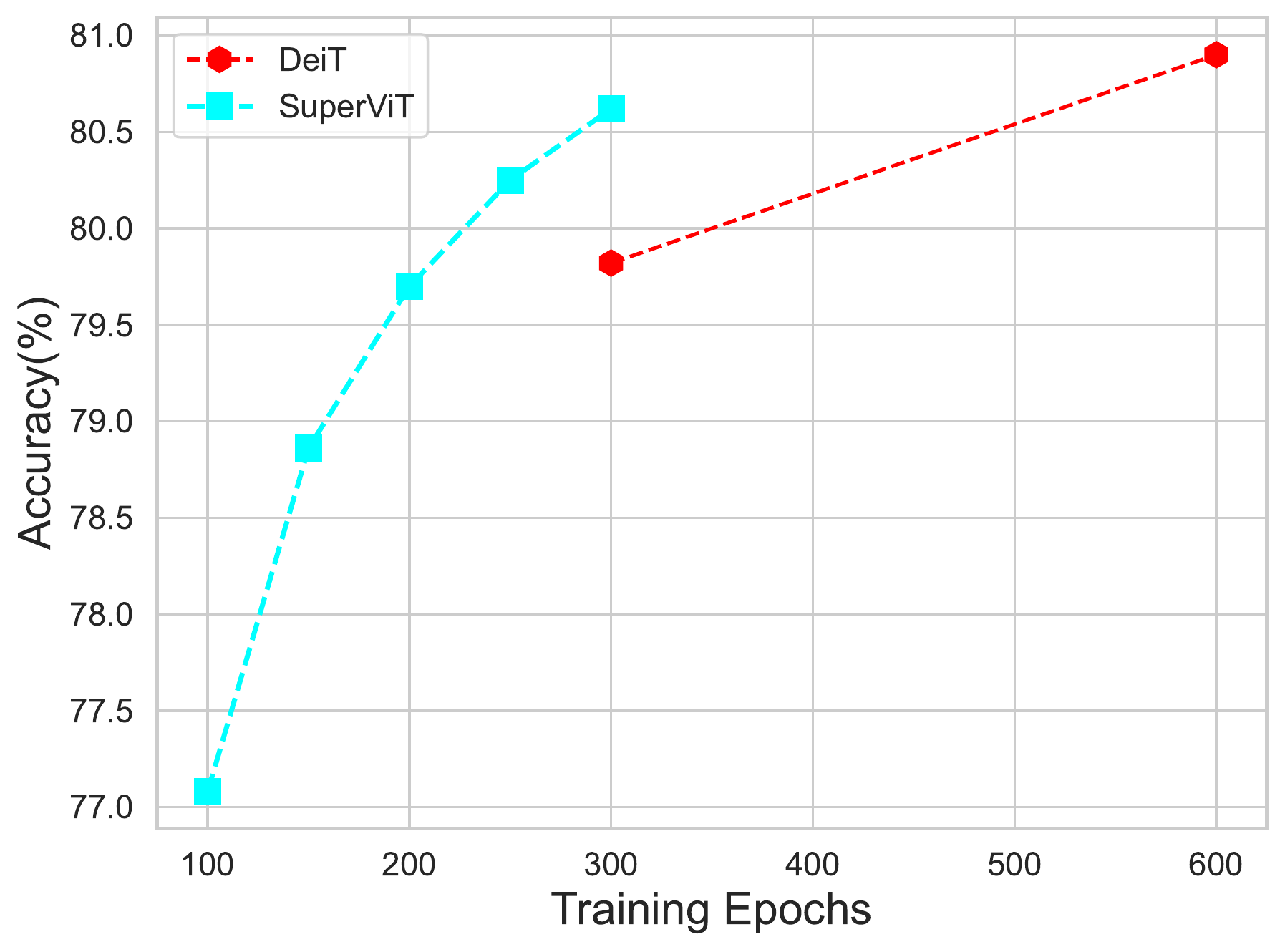}
\end{center}
\caption{\label{training_epoch}Performance of our SuperViT (sequence length: 14$\times$14, keeping rate: 1.0) at different training epochs, and full-trained DeiT-S for 300 and 600 epochs (sequence length: 14$\times$14, keeping rate: 1.0). Experiments are conducted on ImageNet.
}
\end{figure}

\textbf{Comparison with Individual ViT Training}.
Recall that the proposed SuperViT requires three additional forward passes for each training iteration, which seems to increase the training time at each training iteration. Considering this, we perform experiments by training models with different training epochs.
Figure\,\ref{training_epoch} manifests the performance where we provide a total of 100, 150, 200, 250, 300 training epochs for SuperViT, and 300, 600 training epochs for DeiT-S. General speaking, the training time of 150 epochs for SuperViT is almost the same as that of 300 epochs for DeiT.
In this case, the accuracy of SuperViT is 78.86\% while DeiT obtains 79.82\%. It seems that SuperViT has failed to surpass DeiT.
%

%

Nevertheless, we argue that such a comparison is unfair.
As stated in Sec.\,\ref{details}, the sequence length in our multi-size splitting includes \{8$\times$8, 10$\times$10, 12$\times$12, 14$\times$14\} and the multi-token keeping rate includes \{1.0, 0.7, 0.5\}, resulting in a total of 12 different computational costs.
Our training paradigm sequentially forwards one of these 12 cases, which leads to only one SuperViT model that provides improved image recognition performance of 12 subnets with various computational costs.
Therefore, we can make a save conclusion that a more fair comparison should be constructed by training 12 individual ViT models of different complexities, reporting the accuracy performance of each model, and comparing overall training time costs.

\begin{table}[!t]
\begin{center}
\caption{\label{individual_super} Comparison between our training paradigm and individual training ($\dag$). DeiT-S is used as the backbone. Experiments are conducted on ImageNet.}
\begin{tabular}{ccccc}
\hline\noalign{\smallskip}
& {Sequence} & {Keeping} &DeiT-S$^\dag$ &SuperViT \\
 &Length &Rate & Top-1 Acc. & Top-1 Acc.({\color{blue}$\uparrow$}) \\
\hline
& 8$\times$8 & 0.5 &71.6\% &73.9\%{\color{blue}(2.3\%{\color{blue}$\uparrow$})}  \\
& 8$\times$8 & 0.7 &73.5\% &75.3\%{\color{blue}(1.8\%{\color{blue}$\uparrow$})}  \\
& 8$\times$8 & 1.0 &74.8\% & 75.8\%{\color{blue}(1.0\%{\color{blue}$\uparrow$})}   \\
& 10$\times$10 & 0.5 &74.8\% & 77.3\%{\color{blue}(2.5\%{\color{blue}$\uparrow$})}  \\
& 10$\times$10 & 0.7   &76.7\% & 78.3\%{\color{blue}(1.6\%{\color{blue}$\uparrow$})}   \\
& 10$\times$10 & 1.0   &77.4\% & 78.5\%{\color{blue}(1.1\%{\color{blue}$\uparrow$})}   \\
& 12$\times$12 & 0.5   &76.5\% & 78.9\%{\color{blue}(2.4\%{\color{blue}$\uparrow$})}   \\
& 12$\times$12 & 0.7   &78.0\% &79.6\%{\color{blue}(1.6\%{\color{blue}$\uparrow$})}  \\
& 12$\times$12 & 1.0   &78.7\% &79.9\%{\color{blue}(1.2\%{\color{blue}$\uparrow$})} \\
& 14$\times$14 & 0.5   &78.0\% &  80.0\%{\color{blue}($2.0\%${\color{blue}$\uparrow$})}  \\
& 14$\times$14 & 0.7   &79.2\% &  80.5\%{\color{blue}($1.3\%${\color{blue}$\uparrow$})}  \\
& 14$\times$14 & 1.0   &79.8\%  &80.6\%{\color{blue}($0.8\%${\color{blue}$\uparrow$})}  \\
\hline
\end{tabular}
\end{center}
\end{table}

Therefore, we further train each individual subnet from scratch.
Compared with the traditional individual training paradigm, the superiorities of our SuperViT are three folds:
First, significant performance improvement is observed. Table\,\ref{individual_super} shows the accuracy comparison between our SuperViT and its counterpart that trains each ViT model individually. As can be seen, under the same sequence length and keeping rate, our SuperViT increases the performance by 0.8\% $\sim$ 2.5\%. In particular, over 2.0\% performance gains are obtained under a small keeping rate of 0.5. 
Second, our training paradigm is more economical in training consumption. For example, it takes around 4.1 days to train our SuperViT on a single workstation with 4 NVIDIA A100 GPUs when using DeiT-S~\citep{touvron2021training} as the backbone while about 24.4 days are required for traditional individual training paradigm.
Third, our SuperViT is more hardware efficient. As widely discussed in Sec.\,\ref{training_objective} of the main paper, SuperViT can be executable on various hardware platforms equipped with different resources since it can recognize images well at different computational costs. Also, SuperViT empowers instant and adaptive accuracy-efficiency trade-offs by simply adjusting the image patch size and token keeping rate as soon as the available resource on the same platform changes while traditional methods have to carefully download an appropriate model and offload the existing one, which unfortunately lead to many additional data reading overloads in I/O swapping.

\textbf{Transferability}. We further provide performance on downstream tasks including classification on CIFAR-100 dataset~\citep{krizhevsky2009learning} and segmentation on ADE20K dataset~\citep{zhou2017scene}. These experiments are performed and compared using ImageNet pre-trained DeiT-S and our SuperViT.
Table\,\ref{transferability} provides classification results on CIFAR-100. Compared with DeiT-S, the proposed SuperViT serves as a better pre-trained for the downstream classification, leading to performance increase from 88.55\% to 89.33\%.
Table\,\ref{transferability} also gives segmentation results on ADE20K. We observe consistent performance increase even for segmentation task, that is, SuperViT well increases the mIoU results from 42.96 of DeiT-S to 45.32. These results therefore well demonstrate that the transferability of our SuperViT is also superior to that of vanilla DeiT-S, and SuperViT is more appropriate for the downstream tasks.

\begin{table}[!t]
\centering
\caption{Transferring ImageNet pre-trained DeiT-S/SuperViT to CIFAR-100 for classification and ADE20K using UperNet~\citep{xiao2018unified} for segmentation. \label{transferability}}
\resizebox{0.45\textwidth}{!}{
\begin{tabular}{ccc}
\hline
Dataset&CIFAR-100     &ADE20K \\ \hline
Model& Top-1 Acc.(\%) &mIoU   \\ \hline
DeiT-S   & 88.55         &42.96  \\ 
SuperViT &89.33         & 45.32  \\ \hline
\end{tabular}
}
\end{table}

\begin{table*}[!t]
\begin{center}
\caption{\label{token_type}Subnet performance of removing different token types on ImageNet.
DeiT-S is used as the backbone with Sequence Length: $14\times14$, Keeping Rate: $1.0$, Top-1 Acc.: 79.8\%, FLOPs: 4.6G and Throughput: 2461 img./s.
}
\resizebox{\textwidth}{!}{
\begin{tabular}{cccccc}
\hline\noalign{\smallskip}
\multirow{2}{*}{Token Type} & {Sequence} &Keeping &Top-1 & FLOPs{\color{blue}$\downarrow$} &Throughput{\color{blue}$\uparrow$}\\
 &Length &Rate &Acc.{\color{blue}$\uparrow$} (\%) & (G) &(img./s) \\
\hline
\multirow{12}{*}{All Low-attended}
& 14$\times$14 &0.5 &80.0{\color{blue}($+0.2$)} &2.3{\color{blue}($\downarrow$50\%)} & 4767{\color{blue}($\uparrow$1.94$\times$)} \\
& 14$\times$14 & 0.7 & 80.5{\color{blue}($+0.7$)} & 3.0{\color{blue}($\downarrow$35\%)} &3654{\color{blue}($\uparrow$1.48$\times$)} \\
&14$\times$14 &1.0 &80.6{\color{blue}($+0.8$)} &4.6{\color{blue}($\downarrow$0\%)} & 2461{\color{blue}($\uparrow$1.00$\times$)} \\
 &12$\times$12 & 0.5 &78.9{\color{blue}($-0.9$)}  &1.7{\color{blue}($\downarrow$64\%)} &6308{\color{blue}($\uparrow$2.56$\times$)} \\
 &12$\times$12 & 0.7 &79.6{\color{blue}($-0.2$)} &2.2{\color{blue}($\downarrow$52\%)} & 4996{\color{blue}($\uparrow$2.03$\times$)} \\
 &12$\times$12 &1.0 &79.9{\color{blue}($+0.1$)} &3.3{\color{blue}($\downarrow$28\%)} & 3371{\color{blue}($\uparrow$1.37$\times$)} \\
 &10$\times$10 &0.5 &77.3{\color{blue}($-2.5$)}  & 1.2{\color{blue}($\downarrow$74\%)} &9657{\color{blue}($\uparrow$3.88$\times$)} \\
 &10$\times$10 &0.7 & 78.3{\color{blue}($-1.5$)}  &1.5{\color{blue}($\downarrow$68\%)} &7567{\color{blue}($\uparrow$3.01$\times$)} \\
 &10$\times$10 &1.0 &78.5{\color{blue}($-1.3$)} &2.3{\color{blue}($\downarrow$50\%)} & 5173{\color{blue}($\uparrow$2.10$\times$)} \\
 & 8$\times$8 & 0.5 &73.9{\color{blue}($-5.9$)}  &0.7{\color{blue}($\downarrow$85\%)}  &13548{\color{blue}($\uparrow$5.50$\times$)}  \\
 &8$\times$8 &0.7 & 75.3{\color{blue}($-4.5$)}  &1.0{\color{blue}($\downarrow$79\%)} & 10669{\color{blue}($\uparrow$4.33$\times$)} \\
 &8$\times$8 & 1.0 & 75.8{\color{blue}($-4.0$)} & 1.4{\color{blue}($\downarrow$70\%)} &7727{\color{blue}($\uparrow$3.13$\times$)} \\
\hline

&14$\times$14 & 0.5 & 78.9{\color{blue}($-0.9$)} & 2.3{\color{blue}($\downarrow$50\%)} &4767{\color{blue}($\uparrow$1.94$\times$)} \\
&14$\times$14 &0.7 &79.2{\color{blue}($-0.6$)} &3.0{\color{blue}($\downarrow$35\%)} & 3654{\color{blue}($\uparrow$1.48$\times$)} \\
&14$\times$14 &1.0 & 80.1{\color{blue}($+0.3$)} & 4.6{\color{blue}($\downarrow$0\%)} & 2461{\color{blue}($\uparrow$1.00$\times$)} \\
 & 12$\times$12 &0.5 &77.6{\color{blue}($-2.2$)}  & 1.7{\color{blue}($\downarrow$64\%)} & 6308{\color{blue}($\uparrow$2.56$\times$)} \\
 &12$\times$12 & 0.7 & 78.1{\color{blue}($-1.7$)} &2.2{\color{blue}($\downarrow$52\%)} & 4996{\color{blue}($\uparrow$2.03$\times$)} \\
 20\% High-attended & 12$\times$12 &1.0 & 79.0{\color{blue}($-0.8$)} & 3.3{\color{blue}($\downarrow$28\%)} &  3371{\color{blue}($\uparrow$1.37$\times$)} \\
 80\% Low-attended & 10$\times$10 & 0.5 & 75.9{\color{blue}($-3.9$)}  & 1.2{\color{blue}($\downarrow$74\%)} & 9657{\color{blue}($\uparrow$3.88$\times$)} \\
 & 10$\times$10 & 0.7 & 76.6{\color{blue}($-3.2$)}  & 1.5{\color{blue}($\downarrow$68\%)} & 7567{\color{blue}($\uparrow$3.01$\times$)} \\
 & 10$\times$10 & 1.0 & 77.7{\color{blue}($-2.1$)}  & 2.3{\color{blue}($\downarrow$50\%)} &  5173{\color{blue}($\uparrow$2.10$\times$)} \\
 & 8$\times$8 & 0.5 & 72.6{\color{blue}($-7.2$)}  & 0.7{\color{blue}($\downarrow$85\%)}  &  13548{\color{blue}($\uparrow$5.50$\times$)}  \\
 & 8$\times$8 & 0.7 & 73.5{\color{blue}($-6.3$)}  & 1.0{\color{blue}($\downarrow$79\%)} & 10669{\color{blue}($\uparrow$4.33$\times$)} \\
 & 8$\times$8 & 1.0 & 74.9{\color{blue}($-4.9$)} & 1.4{\color{blue}($\downarrow$70\%)} & 7727{\color{blue}($\uparrow$3.13$\times$)} \\
\hline
\end{tabular}
}
\end{center}
\end{table*}

\textbf{Token Type}.
Recall we use the class token as an indicator to reflect the information richness of each token, which is also adopted in many existing studies~\citep{liang2022evit,xu2022evovit,chen2022coarse}. Then, we drop/preserve tokens with lower/higher values of attention. Herein, we try a different dropping strategy by removing 80\% lower-attended regions, but also 20\% high-attended regions. Table\,\ref{token_type} compares the subnet performance of removing different token types. Not surprisingly, removing high-attended tokens causes a very poor performance, \emph{e.g.}, 78.9\% \emph{v.s.} 80.0\% for sequence length of 14$\times$14 and keeping rate of 0.5. These results confirm the correctness of using attention as an information richness indicator.

\textbf{Dropping Blocks}. As stated in Sec.\,\ref{details}, for fair comparison, we follow existing studies to remove less informative tokens at the 4-th, 7-th and 10-th blocks for both DeiT and LV-ViT. Herein, we conduct two other options for ablation by removing tokens at blocks 3/6/9  and blocks 4/6/8. Table\,\ref{dropping_layer} manifests the performance of all subnets \emph{w.r.t.} the three scenarios of brock dropping. In general, we observe the best performance when conducting drops at the 4-th, 7-th and 10-th blocks, in particular to cases of smaller sequence length.
For example, with sequence length of 8$\times$8 and keeping rate of 1.0, ``4/7/10'' obtains 75.8\%, while ``3/6/9'' has 75.5\% and ``4/6/8'' drops a lot to 74.9\%. More performance drops for ``3/6/9'' are attributed to inaccurate attention scores in the shallower third layers. Besides, the redundancy accumulates block-by-block. It would be better to remove tokens of blocks from which the redundancy arises a lot. As a consequence, the results of ``4/6/8'' performs much poor than these of ``4/7/10''.

\textbf{Performance on Smaller ImageNet}. In Table\,\ref{proportions}, we train on different proportions of ImageNet to ablate the effect of our SuperViT on smaller datasets. First, similar to the full training dataset (100\%), we observe consistently better results than the backbone on much smaller training datasets (50\% and 20\%). Second, the performance increase is much more obvious if trained on smaller dataset. For example, the accuracy gain of the subnet with sequence length of 14$\times$14 and keeping rate of 1.0 is 0.8\% for 100\% training dataset, 1.1\% for 50\%, while it is 3.0\% for 20\%. This is because, compared to the superior performance on 100\% of the dataset (79.8\%), the backbone DeiT-S shows very poor performance on 50\% (73.6\%) and 20\% (55.7\%) of the training dataset, upon which a further improvement is much easier.

\textbf{Patch Dimension Alignment}.
Remember in Sec.\,\ref{multi-size-patch} we choose the economical bilinear interpolation to downsample/upsample different local patches for a shape alignment. In Table\,\ref{dimension_align}, we further compare the bilinear interpolation with other alternatives of individual patch embedding layers and bicubic interpolation. We observe that the adopted bilinear interpolation performs the best among all. It is worth discussing that adding individual patch embedding layers results in the worst performance, even though additional $1.83$M parameters are introduced. Compared to the individual embedding, the interpolation methods share one embedding layer, which not only saves more parameters, but also enhances the ability to process multi-scale inputs. Consequently, the shared embedding outstands the individual embedding.

\begin{table*}[!t]
\begin{center}
\caption{\label{dropping_layer}Subnet performance of different block dropping scenarios on ImageNet.
DeiT-S is used as the backbone with Sequence Length: $14\times14$, Keeping Rate: $1.0$, Top-1 Acc.: 79.8\%, FLOPs: 4.6G and Throughput: 2461 img./s.
}
\resizebox{\textwidth}{!}{
\begin{tabular}{cccccc}
\hline\noalign{\smallskip}
\multirow{2}{*}{ Dropping Blocks} & Sequence & Keeping &  Top-1 & FLOPs{\color{blue}$\downarrow$} & Throughput{\color{blue}$\uparrow$}\\
 & Length & Rate & Acc.{\color{blue}$\uparrow$} (\%) & (G) & (img./s) \\
\hline
\multirow{12}{*}{ 4/7/10}
&  14$\times$14 &  0.5 &   80.0{\color{blue}($+0.2$)} &  2.3{\color{blue}($\downarrow$50\%)} &  4767{\color{blue}($\uparrow$1.94$\times$)} \\
&  14$\times$14 &  0.7 &   80.5{\color{blue}($+0.7$)} &  3.0{\color{blue}($\downarrow$35\%)} &  3654{\color{blue}($\uparrow$1.48$\times$)}\\
&  14$\times$14 &  1.0 &   80.6{\color{blue}($+0.8$)} &  4.6{\color{blue}($\downarrow$0\%)} &  2461{\color{blue}($\uparrow$1.00$\times$)} \\
 &  12$\times$12 &  0.5 &  78.9{\color{blue}($-0.9$)}  &  1.7{\color{blue}($\downarrow$64\%)} &   6308{\color{blue}($\uparrow$2.56$\times$)} \\
 &  12$\times$12 &  0.7 &   79.6{\color{blue}($-0.2$)} &  2.2{\color{blue}($\downarrow$52\%)} &  4996{\color{blue}($\uparrow$2.03$\times$)} \\
 &  12$\times$12 &  1.0 &   79.9{\color{blue}($+0.1$)} &  3.3{\color{blue}($\downarrow$28\%)} &   3371{\color{blue}($\uparrow$1.37$\times$)} \\
 &  10$\times$10 &  0.5 &  77.3{\color{blue}($-2.5$)}  &  1.2{\color{blue}($\downarrow$74\%)} &  9657{\color{blue}($\uparrow$3.88$\times$)} \\
 &  10$\times$10 &  0.7 &  78.3{\color{blue}($-1.5$)}  &  1.5{\color{blue}($\downarrow$68\%)} &  7567{\color{blue}($\uparrow$3.01$\times$)} \\
 &  10$\times$10 &  1.0 &  78.5{\color{blue}($-1.3$)}  &  2.3{\color{blue}($\downarrow$50\%)} &   5173{\color{blue}($\uparrow$2.10$\times$)} \\
 &  8$\times$8 &  0.5 &  73.9{\color{blue}($-5.9$)}  &  0.7{\color{blue}($\downarrow$85\%)} &   13548{\color{blue}($\uparrow$5.50$\times$)}  \\
 &  8$\times$8 &  0.7 &  75.3{\color{blue}($-4.5$)}  &  1.0{\color{blue}($\downarrow$79\%)} &  10669{\color{blue}($\uparrow$4.33$\times$)} \\
 &  8$\times$8 &  1.0 &  75.8{\color{blue}($-4.0$)} &  1.4{\color{blue}($\downarrow$70\%)} &  7727{\color{blue}($\uparrow$3.13$\times$)} \\
\hline
\multirow{12}{*}{ 3/6/9}
&  14$\times$14 &  0.5 &   79.3{\color{blue}($-0.5$)} &  2.0{\color{blue}($\downarrow$57\%)} &  5398{\color{blue}($\uparrow$2.19$\times$)} \\
&  14$\times$14 &  0.7 &   80.1{\color{blue}($+0.3$)} &  2.8{\color{blue}($\downarrow$40\%)} &  3910{\color{blue}($\uparrow$1.58$\times$)} \\
&  14$\times$14 &  1.0 &   80.5{\color{blue}($+0.7$)} &  4.6{\color{blue}($\downarrow$0\%)} &  2461{\color{blue}($\uparrow$1.00$\times$)} \\
 &  12$\times$12 &  0.5 &  78.0{\color{blue}($-1.8$)} &  1.4{\color{blue}($\downarrow$70\%)} &  7489{\color{blue}($\uparrow$3.04$\times$)} \\
 &  12$\times$12 &  0.7 &   79.1{\color{blue}($-0.8$)} &  2.0{\color{blue}($\downarrow$57\%)} &  5423{\color{blue}($\uparrow$2.20$\times$)} \\
 &  12$\times$12 &  1.0 &   79.6{\color{blue}($-0.2$)} &  3.3{\color{blue}($\downarrow$28\%)} &  3371{\color{blue}($\uparrow$1.37$\times$)} \\
 &  10$\times$10 &  0.5 &  75.9{\color{blue}($-3.9$)} &  1.0{\color{blue}($\downarrow$79\%)} &  11324{\color{blue}($\uparrow$4.60$\times$)} \\
 &  10$\times$10 &  0.7 &  77.7{\color{blue}($-2.1$)} &  1.4{\color{blue}($\downarrow$70\%)} &  8213{\color{blue}($\uparrow$3.33$\times$)} \\
 &  10$\times$10 &  1.0 &  78.2{\color{blue}($-1.6$)} &  2.3{\color{blue}($\downarrow$52\%)} &  5173{\color{blue}($\uparrow$2.10$\times$)} \\
 &  8$\times$8 &  0.5&  72.0{\color{blue}($-7.8$)} &  0.6{\color{blue}($\downarrow$87\%)} &  15102{\color{blue}($\uparrow$6.13$\times$)} \\
 &  8$\times$8 &  0.7 &  74.6{\color{blue}($-5.2$)} &  0.9{\color{blue}($\downarrow$81\%)} &  11679{\color{blue}($\uparrow$4.74$\times$)} \\
 &  8$\times$8 &  1.0&  75.5{\color{blue}($-4.3$)} &  1.4{\color{blue}($\downarrow$70\%)} &  7727{\color{blue}($\uparrow$3.13$\times$)} \\
\hline
\multirow{12}{*}{ 4/6/8}
&  14$\times$14 &  0.5 &   79.0{\color{blue}($-0.8$)} &  2.1{\color{blue}($\downarrow$55\%)} &  5479{\color{blue}($\uparrow$2.18$\times$)} \\
&  14$\times$14 &  0.7 &   79.9{\color{blue}($+0.1$)} &  2.8{\color{blue}($\downarrow$40\%)} &  3890{\color{blue}($\uparrow$1.58$\times$)} \\
&  14$\times$14 &  1.0 &   80.2{\color{blue}($+0.4$)} &  4.6{\color{blue}($\downarrow$0\%)} &  2461{\color{blue}($\uparrow$1.00$\times$)} \\
 &  12$\times$12 &  0.5 &  77.8{\color{blue}($-2.0$)} &  1.5{\color{blue}($\downarrow$68\%)} &  7612{\color{blue}($\uparrow$3.09$\times$)} \\
 &  12$\times$12 &  0.7&   78.8{\color{blue}($-1.0$)} &  2.1{\color{blue}($\downarrow$55\%)} &  5190{\color{blue}($\uparrow$2.10$\times$)}\\
 &  12$\times$12 &  1.0 &   79.1{\color{blue}($-0.7$)} &  3.3{\color{blue}($\downarrow$28\%)} &  3371{\color{blue}($\uparrow$1.37$\times$)} \\
 &  10$\times$10 &  0.5 &  75.9{\color{blue}($-4.0$)} &  1.1{\color{blue}($\downarrow$76\%)} &  10482{\color{blue}($\uparrow$4.26$\times$)} \\
 &  10$\times$10&  0.7&  77.5{\color{blue}($-2.3$)} &  1.4{\color{blue}($\downarrow$70\%)} &  8110{\color{blue}($\uparrow$3.29$\times$)} \\
 &  10$\times$10 &  1.0 &  77.8{\color{blue}($-2.0$)} &  2.3{\color{blue}($\downarrow$50\%)} &  5173{\color{blue}($\uparrow$2.10$\times$)} \\
 &  8$\times$8&  0.5 &  72.0{\color{blue}($-7.8$)} &  0.7{\color{blue}($\downarrow$85\%)} &  13781{\color{blue}($\uparrow$5.60$\times$)} \\
 &  8$\times$8 &  0.7 &  74.3{\color{blue}($-5.5$)} &  0.9{\color{blue}($\downarrow$81\%)} &  11554{\color{blue}($\uparrow$4.69$\times$)} \\
 &  8$\times$8 &  1.0 &  74.9{\color{blue}($-4.9$)} &  1.4{\color{blue}($\downarrow$70\%)} &  7727{\color{blue}($\uparrow$3.13$\times$)} \\
\hline
\end{tabular}
}
\end{center}
\end{table*}

\begin{table*}[!th]
\begin{center}
\caption{\label{proportions}Subnet performance of SuperViT trained on different proportions of the ImageNet.
DeiT-S is used as the backbone with its performance displayed above the dash lines.
}
\resizebox{\textwidth}{!}{
\footnotesize
\begin{tabular}{cccccc}
\hline\noalign{\smallskip}
\multirow{2}{*}{ Proportion} & { Sequence} &  Keeping &   Top-1 &  FLOPs{\color{blue}$\downarrow$} &  Throughput{\color{blue}$\uparrow$}\\
 & Length & Rate &  Acc.{\color{blue}$\uparrow$} (\%) &  (G) & (img./s) \\
\hline
\multirow{12}{*}{ 100\%}
  &  14$\times$14 & 1.0 & 79.8 & 4.6 &  2461\\\cdashline{2-6}
&  14$\times$14 &  0.5 &   80.0{\color{blue}($+0.2$)} &  2.3{\color{blue}($\downarrow$50\%)} &  4767{\color{blue}($\uparrow$1.94$\times$)} \\
&  14$\times$14 &  0.7 &   80.5{\color{blue}($+0.7$)} &  3.0{\color{blue}($\downarrow$35\%)} &  3654{\color{blue}($\uparrow$1.48$\times$)} \\
&  14$\times$14 &  1.0 &   80.6{\color{blue}($+0.8$)} &  4.6{\color{blue}($\downarrow$0\%)} &  2461{\color{blue}($\uparrow$1.00$\times$)} \\
 &  12$\times$12 &  0.5 &  78.9{\color{blue}($-0.9$)}  &  1.7{\color{blue}($\downarrow$64\%)} &   6308{\color{blue}($\uparrow$2.56$\times$)} \\
 &  12$\times$12 &  0.7 &   79.6{\color{blue}($-0.2$)} &  2.2{\color{blue}($\downarrow$52\%)} &  4996{\color{blue}($\uparrow$2.03$\times$)} \\
 &  12$\times$12 &  1.0 &   79.9{\color{blue}($+0.1$)} &  3.3{\color{blue}($\downarrow$28\%)} &   3371{\color{blue}($\uparrow$1.37$\times$)} \\
 &  10$\times$10 &  0.5 &  77.3{\color{blue}($-2.5$)}  &  1.2{\color{blue}($\downarrow$74\%)} &  9657{\color{blue}($\uparrow$3.88$\times$)} \\
 &  10$\times$10 &  0.7 &  78.3{\color{blue}($-1.5$)}  &  1.5{\color{blue}($\downarrow$68\%)} &  7567{\color{blue}($\uparrow$3.01$\times$)} \\
 &  10$\times$10 &  1.0 &  78.5{\color{blue}($-1.3$)}  &  2.3{\color{blue}($\downarrow$50\%)} &   5173{\color{blue}($\uparrow$2.10$\times$)} \\
 &  8$\times$8 &  0.5 &  73.9{\color{blue}($-5.9$)}  &  0.7{\color{blue}($\downarrow$85\%)}  &   13548{\color{blue}($\uparrow$5.50$\times$)}  \\
 &  8$\times$8 &  0.7 &  75.3{\color{blue}($-4.5$)}  &  1.0{\color{blue}($\downarrow$79\%)} &  10669{\color{blue}($\uparrow$4.33$\times$)} \\
 &  8$\times$8 &  1.0 &  75.8{\color{blue}($-4.0$)} &  1.4{\color{blue}($\downarrow$70\%)} &  7727{\color{blue}($\uparrow$3.13$\times$)} \\
\hline
\multirow{12}{*}{ 50\%}
&  14$\times$14 &  1.0 &   73.6 &  4.6 &  2461 \\\cdashline{2-6}
&  14$\times$14 &  0.5 &   74.3{\color{blue}($+0.7$)} &  2.3{\color{blue}($\downarrow$50\%)} &  4767{\color{blue}($\uparrow$1.94$\times$)} \\
&  14$\times$14 &  0.7 &   74.8{\color{blue}($+1.2$)} &  3.0{\color{blue}($\downarrow$35\%)} &  3654{\color{blue}($\uparrow$1.48$\times$)} \\
&  14$\times$14 &  1.0 &   74.7{\color{blue}($+1.1$)} &  4.6{\color{blue}($\downarrow$0\%)} &  2461{\color{blue}($\uparrow$1.00$\times$)} \\
 &  12$\times$12 &  0.5 &  72.9{\color{blue}($-0.7$)} &  1.7{\color{blue}($\downarrow$64\%)} &  6308{\color{blue}($\uparrow$2.56$\times$)} \\
 &  12$\times$12 &  0.7 &   73.8{\color{blue}($+0.2$)} &  2.2{\color{blue}($\downarrow$52\%)} &  4996{\color{blue}($\uparrow$2.03$\times$)} \\
 &  12$\times$12 &  1.0 &   73.7{\color{blue}($+0.1$)} &  3.3{\color{blue}($\downarrow$28\%)} &  3371{\color{blue}($\uparrow$1.37$\times$)} \\
 &  10$\times$10 &  0.5 &  71.1{\color{blue}($-2.5$)} &  1.2{\color{blue}($\downarrow$74\%)} &  9657{\color{blue}($\uparrow$3.88$\times$)} \\
 &  10$\times$10 &  0.7 &  72.1{\color{blue}($-1.5$)} &  1.5{\color{blue}($\downarrow$68\%)} &  7567{\color{blue}($\uparrow$3.01$\times$)} \\
 &  10$\times$10 &  1.0 &  72.1{\color{blue}($-1.5$)} &  2.3{\color{blue}($\downarrow$50\%)} &  5173{\color{blue}($\uparrow$2.10$\times$)} \\
 &  8$\times$8 &  0.5 &  67.6{\color{blue}($-6.0$)} &  0.7{\color{blue}($\downarrow$85\%)} &  13548{\color{blue}($\uparrow$5.50$\times$)} \\
 &  8$\times$8 &  0.7 &  69.4{\color{blue}($-4.2$)} &  1.0{\color{blue}($\downarrow$79\%)} &  10669{\color{blue}($\uparrow$4.33$\times$)} \\
 &  8$\times$8 &  1.0 &  69.9{\color{blue}($-3.7$)} &  1.4{\color{blue}($\downarrow$70\%)} &  7727{\color{blue}($\uparrow$3.13$\times$)} \\
\hline
\multirow{12}{*}{ 20\%}
&  14$\times$14 &  1.0 &   55.7 &  4.6 &  2461 \\\cdashline{2-6}
&  14$\times$14 &  0.5 &   58.0{\color{blue}($+2.3$)} &  2.3{\color{blue}($\downarrow$50\%)} &  4767{\color{blue}($\uparrow$1.94$\times$)} \\
&  14$\times$14 &  0.7 &   59.0{\color{blue}($+3.3$)} &  3.0{\color{blue}($\downarrow$35\%)} &  3654{\color{blue}($\uparrow$1.48$\times$)} \\
&  14$\times$14 &  1.0 &   58.7{\color{blue}($+3.0$)} &  4.6{\color{blue}($\downarrow$0\%)} &  2461{\color{blue}($\uparrow$1.00$\times$)} \\
 &  12$\times$12 &  0.5 &  56.1{\color{blue}($+0.4$)} &  1.7{\color{blue}($\downarrow$64\%)} &  6308{\color{blue}($\uparrow$2.56$\times$)} \\
 &  12$\times$12 &  0.7 &   57.3{\color{blue}($+1.6$)} &  2.2{\color{blue}($\downarrow$52\%)} &  4996{\color{blue}($\uparrow$2.03$\times$)} \\
 &  12$\times$12 &  1.0 &   57.3{\color{blue}($+1.6$)} &  3.3{\color{blue}($\downarrow$28\%)} &  3371{\color{blue}($\uparrow$1.37$\times$)} \\
 &  10$\times$10 &  0.5 &  53.9{\color{blue}($-1.8$)} &  1.2{\color{blue}($\downarrow$74\%)} &  9657{\color{blue}($\uparrow$3.88$\times$)} \\
 &  10$\times$10 &  0.7 &  55.3{\color{blue}($-0.4$)} &  1.5{\color{blue}($\downarrow$68\%)} &  7567{\color{blue}($\uparrow$3.01$\times$)} \\
 &  10$\times$10 &  1.0 &  55.5{\color{blue}($-0.2$)} &  2.3{\color{blue}($\downarrow$50\%)} &  5173{\color{blue}($\uparrow$2.10$\times$)} \\
 &  8$\times$8 &  0.5 &  49.9{\color{blue}($-5.8$)} &  0.7{\color{blue}($\downarrow$85\%)} &  13548{\color{blue}($\uparrow$5.50$\times$)} \\
 &  8$\times$8 &  0.7 &  51.7{\color{blue}($-4.0$)} &  1.0{\color{blue}($\downarrow$79\%)} &  10669{\color{blue}($\uparrow$4.33$\times$)} \\
 &  8$\times$8 &  1.0 &  52.2{\color{blue}($-3.5$)} &  1.4{\color{blue}($\downarrow$70\%)} &  7727{\color{blue}($\uparrow$3.13$\times$)} \\
\hline
\end{tabular}
}
\end{center}
\end{table*}

\begin{table*}[!t]
\begin{center}
\caption{\label{dimension_align} Subnet performance of different patch dimension alignment methods on ImageNet.
DeiT-S is used as the backbone with Sequence Length: $14\times14$, Keeping Rate: $1.0$, Top-1 Acc.: 79.8\%, FLOPs: 4.6G and Throughput: 2461 img./s.
}
\resizebox{\textwidth}{!}{
\begin{tabular}{cccccc}
\hline\noalign{\smallskip}
\multirow{2}{*}{Alignment Methods} & {Sequence} & Keeping &  Top-1 & FLOPs{\color{blue}$\downarrow$} & Throughput{\color{blue}$\uparrow$}\\
 &Length &Rate & Acc.{\color{blue}$\uparrow$} (\%) & (G) &(img./s) \\
\hline
\multirow{12}{*}{Bilinear Interpolation}
& 14$\times$14 & 0.5 &  80.0{\color{blue}($+0.2$)} & 2.3{\color{blue}($\downarrow$50\%)} & 4767{\color{blue}($\uparrow$1.94$\times$)} \\
& 14$\times$14 & 0.7 &  80.5{\color{blue}($+0.7$)} & 3.0{\color{blue}($\downarrow$35\%)} & 3654{\color{blue}($\uparrow$1.48$\times$)} \\
& 14$\times$14 & 1.0 &  80.6{\color{blue}($+0.8$)} & 4.6{\color{blue}($\downarrow$0\%)} & 2461{\color{blue}($\uparrow$1.00$\times$)} \\
 & 12$\times$12 & 0.5 & 78.9{\color{blue}($-0.9$)}  & 1.7{\color{blue}($\downarrow$64\%)} &  6308{\color{blue}($\uparrow$2.56$\times$)} \\
 & 12$\times$12 & 0.7 &  79.6{\color{blue}($-0.2$)} & 2.2{\color{blue}($\downarrow$52\%)} & 4996{\color{blue}($\uparrow$2.03$\times$)} \\
 & 12$\times$12 & 1.0 &  79.9{\color{blue}($+0.1$)} & 3.3{\color{blue}($\downarrow$28\%)} &  3371{\color{blue}($\uparrow$1.37$\times$)} \\
 & 10$\times$10 & 0.5 & 77.3{\color{blue}($-2.5$)}  & 1.2{\color{blue}($\downarrow$74\%)} & 9657{\color{blue}($\uparrow$3.88$\times$)} \\
 & 10$\times$10 & 0.7 & 78.3{\color{blue}($-1.5$)}  & 1.5{\color{blue}($\downarrow$68\%)} & 7567{\color{blue}($\uparrow$3.01$\times$)} \\
 & 10$\times$10 & 1.0 & 78.5{\color{blue}($-1.3$)}  & 2.3{\color{blue}($\downarrow$50\%)} &  5173{\color{blue}($\uparrow$2.10$\times$)} \\
 & 8$\times$8 & 0.5 & 73.9{\color{blue}($-5.9$)}  & 0.7{\color{blue}($\downarrow$85\%)}  &  13548{\color{blue}($\uparrow$5.50$\times$)}  \\
 & 8$\times$8 & 0.7 & 75.3{\color{blue}($-4.5$)}  & 1.0{\color{blue}($\downarrow$79\%)} & 10669{\color{blue}($\uparrow$4.33$\times$)} \\
 & 8$\times$8 & 1.0 & 75.8{\color{blue}($-4.0$)} & 1.4{\color{blue}($\downarrow$70\%)} & 7727{\color{blue}($\uparrow$3.13$\times$)} \\
\hline
\multirow{12}{*}{Bicubic Interpolation}
& 14$\times$14 & 0.5 &  79.6{\color{blue}($-0.2$)} & 2.3{\color{blue}($\downarrow$50\%)} & 4767{\color{blue}($\uparrow$1.94$\times$)} \\
& 14$\times$14 & 0.7 &  80.5{\color{blue}($+0.7$)} & 3.0{\color{blue}($\downarrow$35\%)} & 3654{\color{blue}($\uparrow$1.48$\times$)} \\
& 14$\times$14 & 1.0 &  80.6{\color{blue}($+0.8$)} & 4.6{\color{blue}($\downarrow$0\%)} & 2461{\color{blue}($\uparrow$1.00$\times$)} \\
 & 12$\times$12 & 0.5 & 78.8{\color{blue}($-1.0$)}  & 1.7{\color{blue}($\downarrow$64\%)} &  6308{\color{blue}($\uparrow$2.56$\times$)} \\
 & 12$\times$12 & 0.7 &  79.5{\color{blue}($-0.3$)} & 2.2{\color{blue}($\downarrow$52\%)} & 4996{\color{blue}($\uparrow$2.03$\times$)} \\
 & 12$\times$12 & 1.0 &  79.6{\color{blue}($-0.2$)} & 3.3{\color{blue}($\downarrow$28\%)} &  3371{\color{blue}($\uparrow$1.37$\times$)} \\
 & 10$\times$10 & 0.5 & 77.4{\color{blue}($-2.4$)}  & 1.2{\color{blue}($\downarrow$74\%)} & 9657{\color{blue}($\uparrow$3.88$\times$)} \\
 & 10$\times$10 & 0.7 & 78.2{\color{blue}($-1.6$)}  & 1.5{\color{blue}($\downarrow$68\%)} & 7567{\color{blue}($\uparrow$3.01$\times$)} \\
 & 10$\times$10 & 1.0 & 78.4{\color{blue}($-1.4$)}  & 2.3{\color{blue}($\downarrow$50\%)} &  5173{\color{blue}($\uparrow$2.10$\times$)} \\
 & 8$\times$8 & 0.5 & 73.8{\color{blue}($-6.0$)}  & 0.7{\color{blue}($\downarrow$85\%)}  &  13548{\color{blue}($\uparrow$5.50$\times$)}  \\
 & 8$\times$8 & 0.7 & 75.4{\color{blue}($-4.4$)}  & 1.0{\color{blue}($\downarrow$79\%)} & 10669{\color{blue}($\uparrow$4.33$\times$)} \\
 & 8$\times$8 & 1.0 & 75.8{\color{blue}($-4.0$)} & 1.4{\color{blue}($\downarrow$70\%)} & 7727{\color{blue}($\uparrow$3.13$\times$)} \\
\hline
\multirow{12}{*}{Individual Embedding}
& 14$\times$14 & 0.5 &  79.6{\color{blue}($-0.2$)} & 2.3{\color{blue}($\downarrow$50\%)} & 4767{\color{blue}($\uparrow$1.94$\times$)} \\
& 14$\times$14 & 0.7 &  80.3{\color{blue}($+0.5$)} & 3.0{\color{blue}($\downarrow$35\%)} & 3654{\color{blue}($\uparrow$1.48$\times$)} \\
& 14$\times$14 & 1.0 &  80.5{\color{blue}($+0.7$)} & 4.6{\color{blue}($\downarrow$0\%)} & 2461{\color{blue}($\uparrow$1.00$\times$)} \\
 & 12$\times$12 & 0.5 & 78.4{\color{blue}($-1.4$)}  & 1.7{\color{blue}($\downarrow$64\%)} &  6308{\color{blue}($\uparrow$2.56$\times$)} \\
 & 12$\times$12 & 0.7 &  79.3{\color{blue}($-0.5$)} & 2.2{\color{blue}($\downarrow$52\%)} & 4996{\color{blue}($\uparrow$2.03$\times$)} \\
 & 12$\times$12 & 1.0 &  79.5{\color{blue}($-0.3$)} & 3.3{\color{blue}($\downarrow$28\%)} &  3371{\color{blue}($\uparrow$1.37$\times$)} \\
 & 10$\times$10 & 0.5 & 76.7{\color{blue}($-3.1$)}  & 1.2{\color{blue}($\downarrow$74\%)} & 9657{\color{blue}($\uparrow$3.88$\times$)} \\
 & 10$\times$10 & 0.7 & 78.0{\color{blue}($-1.8$)}  & 1.5{\color{blue}($\downarrow$68\%)} & 7567{\color{blue}($\uparrow$3.01$\times$)} \\
 & 10$\times$10 & 1.0 & 78.3{\color{blue}($-1.5$)}  & 2.3{\color{blue}($\downarrow$50\%)} &  5173{\color{blue}($\uparrow$2.10$\times$)} \\
 & 8$\times$8 & 0.5 & 73.4{\color{blue}($-6.4$)}  & 0.7{\color{blue}($\downarrow$85\%)}  &  13548{\color{blue}($\uparrow$5.50$\times$)}  \\
 & 8$\times$8 & 0.7 & 75.1{\color{blue}($-4.7$)}  & 1.0{\color{blue}($\downarrow$79\%)} & 10669{\color{blue}($\uparrow$4.33$\times$)} \\
 & 8$\times$8 & 1.0 & 75.9{\color{blue}($-3.9$)} & 1.4{\color{blue}($\downarrow$70\%)} & 7727{\color{blue}($\uparrow$3.13$\times$)} \\
\hline
\end{tabular}
}
\end{center}
\end{table*}

\section{Limitations and Future}
We further discuss unexplored limitations, which will be our future focus.
First, following most compared methods
, we verify our SuperViT on the classification task while its efficacy on dense predictions such as detection and segmentation remains unexplored.
Second, we construct SuperViT from two dimensions of image patch sizes and token keeping rates. More efforts can be made to take into account the transformer's depth, width of token embedding and so on.
Lastly, the vanilla DeiT and LV-ViT consist of plain structures where the number of tokens maintains unchanged in their inputs and outputs. More validations are expected to perform on the pyramid structures~\citep{wang2021pyramid,liu2021swin}.

\section{Conclusion}
%
%
Here, 
we have presented a novel training paradigm to reduce the computational costs in vision transformers (ViTs).
We first break down input images into multiple token sequences of different lengths first.
In each training iteration, we sequentially feed each token sequence to the network where multiple token keeping rates are imposed as well to learn category prediction distributions at different complexities.
Consequently, the trained ViT model, referred to as super vision transformer (SuperViT) in this paper, is demonstrated to provide a better image recognition at a computationally more economical manner compared with existing state-of-the-art methods.
Only one SuperViT model is able to process images at different costs thus it also allows efficient hardware utilization in comparison with traditional methods that have to train numerous ViT models in advance.

%

\section*{Declarations}


\begin{itemize}
\item Funding: This work was supported by National Key R\&D Program of China (No. 2022ZD0118202), the National Science Fund for Distinguished Young Scholars (No. 62025603), the National Natural Science Foundation of China (No. U21B2037, No. U22B2051, No. 62176222, No. 62176223, No. 62176226, No. 62072386, No. 62072387, No. 62072389, No. 62002305 and No. 62272401), and the Natural Science Foundation of Fujian Province of China (No. 2021J01002,  No. 2022J06001).
\item Competing interests: The authors declare that the research was conducted in the absence of any commercial or financial relationships that could be construed as a potential conflict of interest.
\item Availability of data and materials: The dataset ImageNet-1k for this study can be downloaded at: \url{https://www.image-net.org/download.php}.
The dataset CIFAR-100 for this study can be downloaded at: \url{https://www.cs.toronto.edu/~kriz/cifar.html}.
The dataset ADE20K for this study can be downloaded at: \url{https://groups.csail.mit.edu/vision/datasets/ADE20K/}.
\item Code availability: Code is made publicly available at     \url{https://github.com/lmbxmu/SuperViT}.  
\item Authors' contributions: 
Material preparation, data collection and analysis were mostly performed by Mingbao Lin, 
Mengzhao Chen and Yuxin Zhang. The SuperViT model was originally proposed by Mingbao 
Lin and Mengzhao Chen, improved by Chunhua Shen. Chunhua Shen also 
made 
efforts to 
revise the manuscript. Rongrong Ji and Liujuan Cao, leaders of this project, delved  into specific 
discussions of the feasibility and polishing manuscript. Liujuan Cao was also involved in partial 
experimental designs and paper revision. The first draft of the manuscript was written by Mingbao Lin and all authors commented on previous versions of the manuscript. All authors read and approved the final manuscript.

\end{itemize}

\begin{appendices}




\end{appendices}


\bibliographystyle{bst/sn-basic}
\bibliography{main} 


\end{document}